\documentclass{article}

\usepackage{microtype}
\usepackage{graphicx}
\usepackage{subcaption}
\usepackage{multirow}
\usepackage{booktabs}
\usepackage{stfloats}
\usepackage{hyperref}

\usepackage[accepted]{icml2026}

\usepackage{amsmath}
\usepackage{amssymb}
\usepackage{mathtools}
\usepackage{amsthm}
\usepackage{longtable}
\usepackage{amssymb} 

\usepackage[capitalize,noabbrev]{cleveref}

\icmltitlerunning{MATA-Former \& SIICU: High-Fidelity ICU Prediction}

\begin{document}

\twocolumn[
    \icmltitle{MATA-Former \& SIICU: Semantic Aware Temporal Alignment for \texorpdfstring{\\}{ }
    High-Fidelity ICU Risk Prediction}
    
    \icmlsetsymbol{equal}{*}
    
    \begin{icmlauthorlist}
    \icmlauthor{Zhichong Zheng}{sch1,sch2,equal} 
    \icmlauthor{Xiaohang Nie}{sch2,sch4,equal} 
    \icmlauthor{Xueqi Wang}{sch3} 
    \icmlauthor{Yuanjin Zhao}{sch1} 
    \icmlauthor{Haitao Zhang}{sch3} 
    \icmlauthor{Yichao Tang}{sch1,sch2} 
    \end{icmlauthorlist}
    
    \icmlaffiliation{sch1}{Tongji University}
    
    \icmlaffiliation{sch2}{Shanghai Innovation Institute}
    
    \icmlaffiliation{sch3}{Department of Critical Care Medicine, Shanghai East Hospital, Tongji University School of Medicine} 
    
    \icmlaffiliation{sch4}{Harbin Institute of Technology}

    \icmlcorrespondingauthor{Haitao Zhang}{2505018@tongji.edu.cn} 
    \icmlcorrespondingauthor{Yichao Tang}{tangyichao@tongji.edu.cn}     
    
    \icmlkeywords{Machine Learning, Clinical Time Series, Transformer}
    
    \vskip 0.3in
]

\printAffiliationsAndNotice{\icmlEqualContribution}

\begin{abstract}
Forecasting evolving clinical risks relies on intrinsic pathological dependencies rather than mere chronological proximity, yet current methods struggle with coarse binary supervision and physical timestamps. To align predictive modeling with clinical logic, we propose the Medical-semantics Aware Time-ALiBi Transformer (MATA-Former), utilizing event semantics to dynamically parameterize attention weights to prioritize causal validity over time lags. Furthermore, we introduce Plateau-Gaussian Soft Labeling (PSL), reformulating binary classification into continuous multi-horizon regression for full-trajectory risk modeling. Evaluated on SIICU---a newly constructed dataset featuring over 506k events with rigorous expert-verified, fine-grained annotations---and the MIMIC-IV dataset, our framework demonstrates superior efficacy and robust generalization in capturing risks from text-intensive, irregular clinical time series.
\end{abstract}

\section{Introduction}
\label{sec:intro}
In the domain of critical care modeling, Clinical Decision Support Systems (CDSS) faces a fundamental asymmetry: the continuous stream of high-frequency structured vitals versus the sporadic, semantic-rich events found in unstructured clinical narratives~\cite{zheng2025, seinen2022}. While standard deep learning paradigms have excelled in processing homogeneous time series, they struggle to harness the ``data deluge'' of the Intensive Care Unit (ICU), where critical predictive signals are often buried in textual logs that are underutilized due to extraction complexities~\cite{cardamone2025}. The inability to effectively align these heterogeneous modalities leads to the loss of subtle pathological precursors, resulting in missed intervention windows and catastrophic patient outcomes.

Current approaches struggle to address this modality imbalance. While structured data is effectively modeled, the integration of unstructured text remains superficial due to implicit modeling constraints~\cite{gumiel2021, raghu2023}. Specifically, mainstream architectures are constrained by the implicit assumption of monotonic information decay, often prioritizing proximal context even when processing long-range dependencies~\cite{shukla2020, zerveas2021, press2021}. However, this assumption creates a significant misalignment with clinical reality, where predictive signals are embedded in complex medical semantics rather than mere temporal proximity. We characterize this complexity through two key observations: (1)~\textit{Intrinsic Temporal Validity}~\cite{cai2018}, where features possess distinct effective lifespans, ranging from immediate acute impacts to long-term pharmacological lags; and (2)~\textit{Query-Dependent Relevance}~\cite{lu2022}, where historical relevance is contingent on the specific risk query, and the target risk's semantics dictate whether predictive signals reside in the recent or distant past. Moreover, standard paradigms typically reduce dynamic disease progression to coarse binary targets, ignoring the continuous granularity of risk evolution. Consequently, strictly coupling physical time distance with discrete labels limits model efficacy on irregular, text-intensive time series.

To bridge this semantic-temporal gap, we propose the Medical-semantics Aware Time-ALiBi Transformer (MATA-Former). By integrating unified clinical embeddings with a semantic-guided temporal attention mechanism, MATA-Former dynamically generates query-specific focus windows. This allows the model to internalize the temporal scope as an intrinsic property, aligning attention weights with the pathological relevance of historical events rather than their physical proximity. Furthermore, to address the limitations of binary labels in capturing progressive deterioration, we introduce Plateau-Gaussian Soft Labeling (PSL). This transforms discrete classification into a continuous multi-window regression, smoothing label noise and enabling the capture of dynamic risk trajectories.

High-fidelity evaluation is currently hindered by the scarcity of fine-grained annotations~\cite{miotto2018}. To address this, we present the Semantic-Integrated Intensive Care Unit (SIICU) dataset, extracted from a real-world Hospital Information System (HIS). Tailored for complex ICU scenarios, SIICU comprises 360 expert-selected key risk labels and over 506,000 clinical events explicitly annotated using these labels. We implemented a strict human-AI collaborative annotation protocol: initial pre-annotation was performed using Large Language Models (LLMs), followed by exhaustive verification by human experts.

Our contributions are summarized as follows:
\begin{itemize}
    \item \textbf{MATA-Former Architecture:} To bridge the semantic-temporal gap, we propose MATA-Former, a unified architecture that harmonizes unstructured clinical text with structured vitals. By parameterizing pathological latencies via semantic-guided attention, it achieves adaptive alignment between medical semantics and physical time dynamics.
    
    \item \textbf{PSL Paradigm Shift:} We introduce PSL, a paradigm shift from coarse binary classification to event-wise continuous regression. Through bidirectional Gaussian mapping, PSL achieves dense supervision throughout the entire ICU clinical trajectory, capturing granular risk dynamics from incubation to recovery.
    
    \item \textbf{SIICU Dataset:} We establish SIICU, a high-fidelity dataset constructed via a rigorous ``LLM-pre-annotation + Human-verification'' protocol. Comprising over 506k expert-annotated events, it addresses the critical scarcity of fine-grained, text-intensive clinical datasets for robust evaluation.
\end{itemize}

\section{Related Work}
\label{sec:related_work}

\textbf{Clinical Time Series Modeling.} Clinical temporal modeling is pivotal for CDSS~\cite{shickel2017}. While early rule-based systems~\cite{knaus1985, legall1984} provided foundational baselines, they failed to capture the non-linear dynamics of disease progression. The advent of deep learning saw the widespread adoption of Recurrent Neural Network variants for longitudinal Electronic Health Records (EHR)~\cite{choi2016, lipton2015, rajkomar2018}. However, these recurrent architectures are plagued by gradient vanishing and memory bottlenecks when modeling long-range dependencies. Recently, the ``pre-training and fine-tuning'' paradigm---exemplified by Med-BERT~\cite{rasmy2021} and BEHRT~\cite{li2020}---has leveraged masked language modeling to learn contextual representations. Despite this, current medical foundation models often treat clinical events as generic tokens, neglecting the complex pathological correlations and semantic weights inherent in medical events, thereby limiting their efficacy in capturing fine-grained risk evolution~\cite{raghu2023, karami2024, zheng2025}.

\textbf{Time-Aware Attention Mechanisms.} To address irregular ICU temporal sampling, prior works utilize hierarchical attention~\cite{luo2020}, timestamp encodings~\cite{kazemi2019, tipirneni2022, devries2025}, or text serialization with positional embeddings~\cite{lee2025}, yet often struggle with continuous intervals due to discrete or implicit ordering. Even polynomial trend modeling in TALE-EHR~\cite{yu2025} lacks dynamic contextual adjustment. MATA-Former overcomes this via query-dependent Laplace distribution bias, replacing fixed weights with adaptive temporal receptive fields driven by event semantics.

\textbf{Medical Semantic Learning.} Mapping heterogeneous events to a unified semantic space is critical for processing unstructured, long-tail medical data~\cite{lehman2023}. Traditional methods relying on discrete encodings often suffer from the curse of dimensionality~\cite{choi2017}. Recent advances confirm that utilizing natural language as a unified interface~\cite{luo2022} allows for the extraction of deep semantics from sparse indicators~\cite{su2025, zhang2025, hegselmann2025}. Leveraging pre-trained LLMs to embed clinical terms has proven effective in enhancing representations for rare conditions~\cite{pang2021, li2020}. We extend this strategy by implementing semantically aware alignment: instead of using semantics solely for representation, we utilize them to parameterize the temporal attention mechanism, ensuring that the model's focus aligns with the pathological validity of each event.

\textbf{Clinical Early Warning Benchmarks.} High-quality benchmarks are the cornerstone of valid clinical modeling. While datasets like MIMIC-III/IV~\cite{johnson2016, johnson2023} and eICU-CRD~\cite{pollard2018} have propelled CDSS development, they exhibit significant limitations regarding annotation paradigms. Reliance on discharge-level ICD codes fails to provide event-wise supervision, implicitly introducing look-ahead bias in real-time prediction tasks~\cite{ancker2017, raghu2023}. Point-wise manual annotation for high-density ICU sequences is prohibitively labor-intensive~\cite{miotto2018}. While recent studies explore LLM-assisted pre-annotation~\cite{huang2024, hu2024}, the absence of rigorous human-in-the-loop verification risks introducing label noise that impairs generalization~\cite{berkhout2025}. Traditional binary labels oversimplify risk deterioration into instantaneous switches, ignoring the continuity ~\cite{hyland2020}. Recent research advocates for soft label mapping~\cite{yeche2023, lu2022} to capture dynamic risk trajectories.

\section{The SIICU Dataset}
\label{sec:dataset}
Derived from the Hospital Information System~(HIS) of a tertiary ICU in Eastern China, Semantic-Integrated Intensive Care Unit~(SIICU) comprises 506,119 longitudinal records from 507 patients, capturing full admission-to-discharge trajectories composed of interleaved unstructured text and structured vitals. The study adheres to strict privacy protocols with all Personally Identifiable Information removed, under Medical Ethics Committee approval.

\subsection{Data Structuring and Serialization}
To mitigate long-sequence computational bottlenecks, we utilize a compression protocol that first consolidates concurrent structured metrics into composite events and subsequently maps all events into 12 discrete categories (e.g., Nursing Notes). The resulting stream, serialized by timestamp, significantly reduces sequence length while preserving granular context.

\subsection{Fine-Grained Risk Annotation Protocol}
Traditional Electronic Health Record (EHR) benchmarks typically rely on static, discharge-level ICD codes, often inferring disease onset from sparse metric fluctuations. These coarse labels lack temporal granularity and inherently introduce \textit{look-ahead bias}, rendering them unsuitable for real-time dynamic early warning. To achieve precise event-level risk labeling without the prohibitive cost of manual annotation, we implemented a rigorous Human-in-the-Loop collaborative framework.

\textbf{Schema Definition.} In collaboration with ICU experts, 360 key clinical risk events (e.g., \textit{Cardiogenic Shock}, \textit{Aortic Dissection}) were defined, spanning 7 organ systems and 49 major categories.

\textbf{LLM-Assisted Pre-Annotation.} We leveraged the multimodal reasoning capabilities of Gemini 2.5 Pro to process patient trajectories and generate candidate risks for each event.

\textbf{Expert Verification.} To ensure clinical validity, senior clinicians performed exhaustive verification and correction of all LLM-generated positive samples, alongside audits of sampled negative instances. These verified labels constitute the final fine-grained Ground Truth.

\begin{figure}[t] 
    \centering
    \begin{subfigure}[b]{0.48\columnwidth} 
        \centering
        \includegraphics[width=\linewidth]{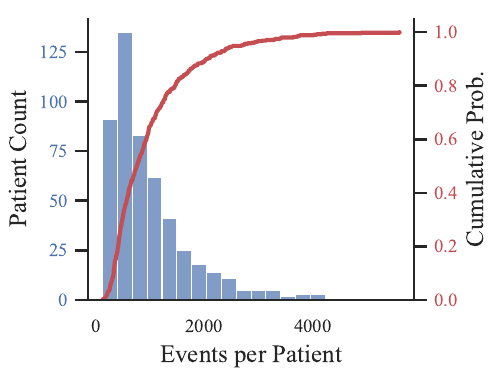}
        \caption{Event Count}
        \label{fig:1a}
    \end{subfigure}
    \hfill 
    \begin{subfigure}[b]{0.48\columnwidth}
        \centering
        \includegraphics[width=\linewidth]{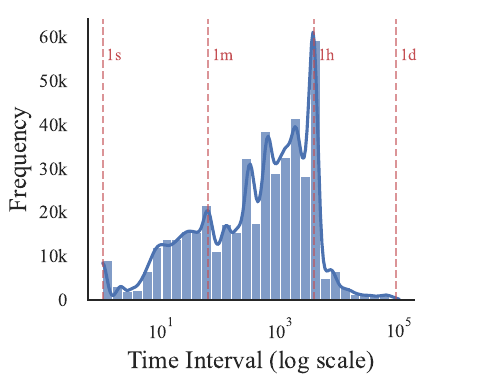}
        \caption{Time Intervals}
        \label{fig:1b}
    \end{subfigure}
    \begin{subfigure}[b]{0.48\columnwidth}
        \centering
        \includegraphics[width=\linewidth]{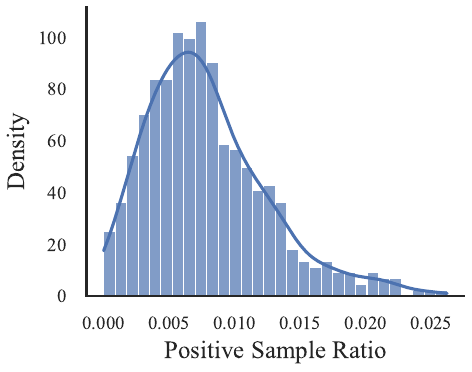}
        \caption{Prob. Density} 
        \label{fig:1c}
    \end{subfigure}
    \hfill
    \begin{subfigure}[b]{0.48\columnwidth}
        \centering
        \includegraphics[width=\linewidth]{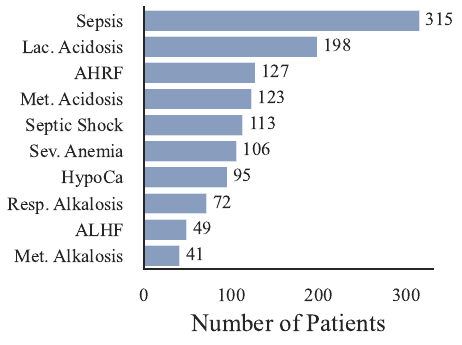}
        \caption{Risk Freq.}
        \label{fig:1d}
    \end{subfigure}
    
    \caption{Statistics of SIICU. (a) Heavy-tailed distribution of sequence lengths. (b) Irregular time intervals. (c) Density of positive sample ratios. (d) Long-tail distribution of risks. These distributions highlight the dataset's high heterogeneity and sparsity.}
    \label{fig:dataset_stats_single_col}
\end{figure}

\begin{figure*}[t]
    \centering
    \includegraphics[width=\linewidth]{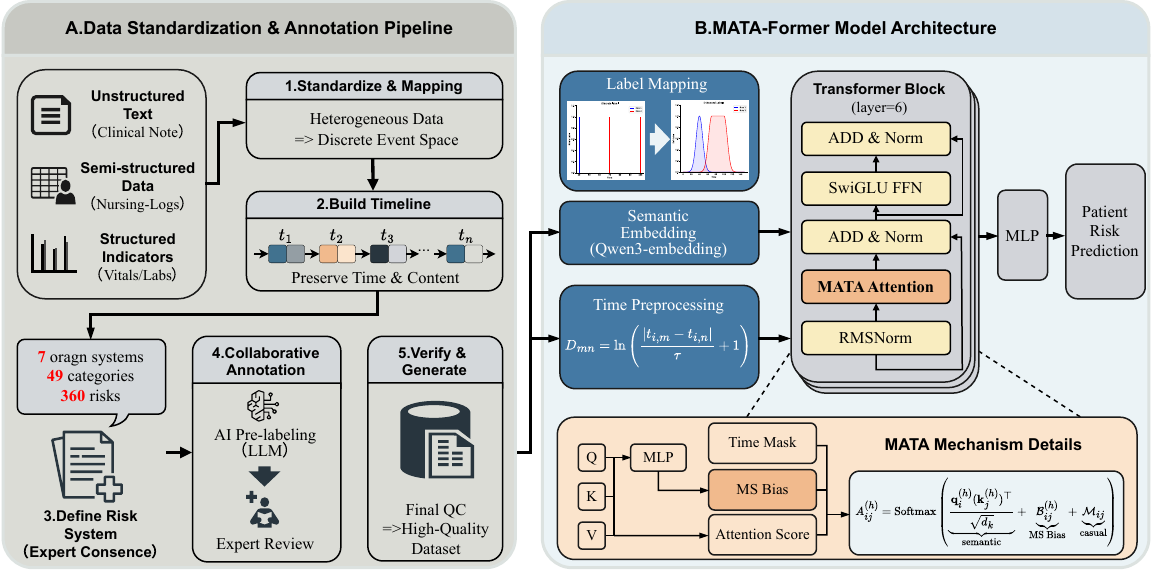}
    \caption{Overview of the SIICU \& MATA-Former framework. (A) The SIICU pipeline transforms heterogeneous clinical records into chronological timelines, utilizing expert-verified collaborative annotation for ground truth generation. (B) The MATA-Former Architecture integrates semantic embeddings with log-transformed temporal features. The core MATA mechanism modulates self-attention via dynamic Laplacian biases, optimized against PSLs for multi-horizon risk prediction.}
    \label{fig:Arch}
\end{figure*}

\subsection{Statistical Characteristics}
SIICU is characterized by high heterogeneity and sparsity. Trajectories follow a heavy-tailed length distribution (\cref{fig:1a}) requiring robustness against attention dilution, while sampling is highly irregular (\cref{fig:1b}) with intervals spanning orders of magnitude---from automated monitoring to sporadic manual interventions. Furthermore, the risk landscape features extreme class imbalance (\cref{fig:1c}) and long-tail distributions (\cref{fig:1d}), where sparse positive samples impede standard optimization and bias models toward dominant head classes.

\section{MATA-Former}
\label{sec:mata_former}

MATA-Former captures complex dependencies in text-intensive ICU streams through the Medical-semantics Aware Time-ALiBi (MATA) module. By augmenting standard Multi-Head Self-Attention (MHSA), this mechanism dynamically modulates attention biases based on query semantics, enabling the generation of adaptive, instance-specific temporal receptive fields.

\subsection{Unified Textualization \& Semantic Embedding}
\label{sec:unified_embedding}

We introduce the Unified Semantic Embedding framework to harmonize heterogeneous clinical data---ranging from unstructured free-text logs to structured vitals---into a uniform textual stream, harnessing the semantic reasoning of pre-trained LLMs to extract latent signals from clinical notes and rare indicators~\cite{su2025, zhang2025}. Formally, unstructured narratives are processed directly as raw text: $Text_{input} = Text_{raw}$. For structured data, to explicitly encode the semantic context of the 12 defined event types (denoted as $C$, e.g., Lab Tests), we serialize the set of $M$ clinical metrics $\{k_m, v_m\}_{m=1}^M$ within a composite event using the following template:

\vspace{-1.4em} 
\begin{equation}
    Text_{input} = \bigoplus_{m=1}^{M} \left( \texttt{[} C \texttt{]} \oplus k_m \oplus \texttt{:} \oplus v_m \right)
\end{equation}

where $\oplus$ denotes string concatenation. This formulation ensures that the event type $C$ serves as a semantic anchor, allowing the model to distinguish between identical numerical values arising from different clinical contexts. The serialized input is encoded via the frozen Qwen3-embedding-8B model ($\mathcal{E}$) as an offline preprocessing step, yielding event-level semantic representations. This approach efficiently compresses sparse, high-dimensional tabular data into dense vectors, mitigating computational redundancy without involving the large encoder in subsequent training.

\begin{equation}
    \mathbf{e} = \mathcal{E}(Text_{input}) \in \mathbb{R}^{4096}
\end{equation}

\subsection{Task Modeling and Problem Definition}
\label{sec:problem_definition}

Let $\mathcal{D} \! = \! \{S_n\}_{n=1}^N$ denote a set of trajectories $S_n = \{(\mathbf{e}_{n,i}, t_{n,i})\}_{i=1}^{L_n}$, comprising $L_n$ event-timestamp pairs where $\mathbf{e}_{n,i} \in \mathbb{R}^{4096}$ is the semantic embedding and $t_{n,i} \in \mathbb{R}^+$ is the physical timestamp. To capture clinical evolution across prospective horizons $\mathcal{K} = \{6\text{h}, 12\text{h}, 24\text{h}, 48\text{h}\}$, we employ Plateau-Gaussian Soft Labeling (PSL) to transform discrete diagnoses into continuous supervision signals.

Formally, let $\mathcal{I}_r = [T_{\text{start}}, T_{\text{end}}]$ denote the active interval for risk $r \in \mathcal{R}$. We model the supervision signal $ y_{i,r,k} \in [0, 1]$ for horizon $k \in \mathcal{K}$ via an unnormalized Gaussian kernel:
\begin{equation}
y_{i,r,k} = \exp\left( -\frac{\delta(t_i, \mathcal{I}_{r})^2}{2\sigma_k^2} \right)
\end{equation}
where $\sigma_k = k$ controls the horizon-specific decay. 
The displacement metric $\delta(t_i, \mathcal{I}_r)$ is defined as:
\begin{equation}
    \delta(t_i, \mathcal{I}_r) = \inf_{x \in \mathcal{I}_r} |t_i - x| = 
    \begin{cases} 
       0 & T_{\text{start}} \le t_i \le T_{\text{end}} \\
       T_{\text{start}} - t_i & t_i < T_{\text{start}} \\
       t_i - T_{\text{end}} & t_i > T_{\text{end}}
    \end{cases}
\end{equation}

This formulation yields a plateau of 1.0 within $\mathcal{I}_r$ with smooth outward decay, effectively modeling gradual risk accumulation and remission. We optimize a mapping $f_\theta: \mathcal{S} \to \mathbb{R}^{|\mathcal{R}| \times |\mathcal{K}|}$ to predict the risk matrix $\hat{\mathbf{Y}}_{i}$ given history $S_{<i}$. Regressing these continuous soft labels captures multi-scale progression dynamics, avoiding the limitations of discrete binary targets.

\subsection{Medical-semantics Aware Time-ALiBi}
\label{sec:ms_aware_attention}
Clinical events exert influence over distinct temporal horizons. To encode this, temporal attention bias is modeled via a dynamic Laplace distribution, where geometric parameters---center shift $\mu$ and decay rate $\alpha$ ---are conditioned on the query $\mathbf{q}$ via $(\mu, \alpha) =\mathcal{F}_\phi(\mathbf{q})$ to achieve adaptive, instance-specific receptive fields. Furthermore, to mitigate heavy-tailed skewness, the time interval between query event $i$ and key event $j$ is log-transformed to a feature space approximating normality, yielding $D_{ij}$:
\begin{equation}
    D_{ij} = \ln\left(\frac{|t_{i} - t_{j}|}{\tau} + 1\right)
\end{equation}
where $\tau$ serves as a scaling factor for temporal granularity.

To capture event-specific dependency patterns within the self-attention mechanism, we integrate a lightweight Multi-Layer Perceptron (MLP) projection module $\mathcal{F}_\phi$. This module dynamically regresses shape parameters from the query $\mathbf{q}_{i}^{(h)}$ for head $h$, yielding query-specific residuals:
\begin{equation}
    \{\delta_{\alpha, i}^{(h)}, \delta_{\mu, i}^{(h)}\} = \mathbf{W}_2 \cdot \text{Tanh}(\mathbf{W}_1 \mathbf{q}_{i}^{(h)} + \mathbf{b}_1)
\end{equation}

We synthesize the final geometric parameters by modulating head-specific priors $\{\bar{\alpha}^{(h)}, \bar{\mu}^{(h)}\}$ with query-specific residuals, fusing global clinical patterns with instance-level context. Specifically, we employ constrained projection functions to update the decay rate $\alpha_{i}^{(h)}$ and center shift $\mu_{i}^{(h)}$, ensuring strictly valid parameter ranges and numerical stability:
\begin{align}
    \alpha_{i}^{(h)} &= \mathcal{P}_{\alpha} \left( \bar{\alpha}^{(h)}, \delta_{\alpha, i}^{(h)} \right) \\
    \mu_{i}^{(h)} &= \mathcal{P}_{\mu} \left( \bar{\mu}^{(h)}, \delta_{\mu, i}^{(h)} \right)
\end{align}
where $\mathcal{P}_{\cdot}$ encapsulates the specific transformation mechanics, as detailed in \cref{mu_alpha_Initialization}.

The resulting query-dependent Laplacian bias $\mathcal{B}^{\text{Time}}$, which encodes the adaptive receptive field, is computed over the log-distance matrix $D \in \mathbb{R}^{L \times L}$:
\begin{equation}
    \mathcal{B}_{ij}^{(h)} = -\alpha_{i}^{(h)} \cdot |D_{ij} - \mu_{i}^{(h)}|
\end{equation}

To prevent look-ahead bias, a causal mask $\mathcal{M}$ governed by physical timestamps $t$ is applied, strictly preserving temporal causality:
\begin{equation}
    \mathcal{M}_{ij} = 
    \begin{cases}  
        0 & \text{if } t_j \le t_i \\ 
        -\infty & \text{if } t_j > t_i 
    \end{cases}
\end{equation}

The final attention weights $\mathbf{A}_{ij}^{(h)}$ are derived by injecting adaptive temporal and causal biases into the semantic scores:
\begin{equation}
    \mathbf{A}_{ij}^{(h)}  = \text{Softmax}\left( \underbrace{\frac{\mathbf{q}_i^{(h)} (\mathbf{k}_j^{(h)})^\top}
    {\sqrt{d_k}}}_{\text{Semantic}} + 
    \underbrace{\mathcal{B}_{i j}^{(h)}}_{\text{MS-Bias}} + 
    \underbrace{\mathcal{M}_{ij}}_{\text{Causal}} \right)
\end{equation}

\subsection{Stability and Prior Initialization}
To mitigate early-stage attention oscillations induced by high variance in the unoptimized projection network $\mathcal{F}_\phi$, we zero-initialize the final layer weights, constraining initial outputs to static bias terms $\{\bar{\alpha}^{(h)}, \bar{\mu}^{(h)}\}$. This approach constructs a multi-scale temporal filter bank via a structured co-initialization strategy. Specifically, peak priors $\bar{\mu}_h$ employ linear tiling across the logarithmic span $[0, \gamma_{\mu}]$ to ensure coverage from instantaneous to long-range history, while slope priors $\bar{\alpha}_h$ anchor to $\alpha_{base}$ with a perturbation $\epsilon$ to break symmetry. The initialization is formalized as:
\begin{equation}
    \begin{aligned} 
        \bar{\mu}^{(h)} &= \frac{h}{H-1} \cdot \gamma_{\mu}, \quad h \in \{0, \dots, H-1\} \\ 
        \bar{\alpha}^{(h)} &= \alpha_{base} + \epsilon, \quad \epsilon \sim \mathcal{U}(-0.05, 0.05) 
    \end{aligned}
\end{equation}
This configuration establishes an initial structured grid over the temporal domain. As training progresses, the model transitions from these static priors to learning dynamic, pathology-specific offsets for refined semantic alignment. Detailed projection logic is provided in \cref{mu_alpha_Initialization}.

\subsection{Optimization Objective}
\label{sec:optimization_objective}

Despite the zero-inflated label distribution, we employ unweighted MSE to avoid noise amplification inherent in weighting mechanisms for continuous soft labels (validated in \cref{sec:ablation_analysis}). Formally, we minimize the loss $\mathcal{L}$ over a mini-batch of $B$ sequences, computed across all risk-horizon dimensions:
\begin{equation} 
\mathcal{L} = \frac{1}{Z} \sum_{n=1}^{B} \sum_{i=1}^{L_n} \sum_{r=1}^{|\mathcal{R}|} \sum_{k=1}^{|\mathcal{K}|} \left( \hat{Y}_{n,i}^{(r,k)} - Y_{n,i}^{(r,k)} \right)^2
\end{equation}
where $Z = |\mathcal{R}| \cdot |\mathcal{K}| \cdot \sum_{n=1}^{B} L_n$ is the normalization factor ensuring scale invariance to variable sequence lengths.

\section{Experiments}
\label{sec:experiments}

\subsection{Experimental Setup and Metrics}
We implement a patient-wise stratified 4-fold cross-validation. To mitigate length-of-stay skew, partitions balance cumulative event counts rather than patient numbers, enforcing strict patient isolation to preclude information leakage. This maintains a consistent 3:1 training-to-testing event distribution: 
\begin{equation}
\sum_{p \in \mathcal{D}_{\text{train}}} L_p \approx 3 \sum_{p \in \mathcal{D}_{\text{test}}} L_p
\end{equation}

To further substantiate the model's generalization capability, we extended our evaluation to the public MIMIC-IV 3.1 benchmark. Specifically, we targeted three high-stakes binary critical care endpoints: Mortality, Sepsis, and Invasive Mechanical Ventilation (IMV) (details in \cref{MIMIC_setup}).

Given the zero-inflated label distribution, evaluation prioritizes Sample-Wise and Micro AUPRC, complemented by AUROC, Precision@K, and Brier Score. Results report the mean and standard deviation across four iterations.

\begin{table*}[t]
\centering
\caption{Comparative performance on the SIICU dataset. Values denote mean $\pm$ standard deviation across four independent runs. Given the extreme label sparsity, evaluation prioritizes AUPRC over AUROC to mitigate the impact of class imbalance. MATA-Former achieving decisive gains in precision-oriented metrics (P@1, P@5) and demonstrating superior sensitivity to rare, high-priority clinical risks.}
\label{tab:SIICU}
\small
\setlength{\tabcolsep}{4pt}
\begin{tabular}{lccccccc}
\toprule
\textbf{Model} & \textbf{AUPRC (Sample)} & \textbf{AUPRC (Micro)} & \textbf{AUROC} & \textbf{Brier Score}  & \textbf{P@1} & \textbf{P@5} \\
\midrule
Doctor AI & $0.205 \pm 0.010$ & $0.187 \pm 0.001$ & $0.960 \pm 0.001$ & $0.064 \pm 0.001$  & $0.331 \pm 0.003$ & $0.286 \pm 0.006$ \\
LSTM & $0.189 \pm 0.017$ & $0.153 \pm 0.027$ & $0.936 \pm 0.017$ & $0.062 \pm 0.006$  & $0.288 \pm 0.033$ & $0.264 \pm 0.025$ \\
Transformer & $0.372 \pm 0.009$ & $0.364 \pm 0.003$ & $0.929 \pm 0.003$  & $0.006 \pm 0.002$ & $0.432 \pm 0.015$ & $0.378 \pm 0.009$ \\
HiTANet & $0.336 \pm 0.014$ & $0.274 \pm 0.010$ & $0.956 \pm 0.002$ & $0.006 \pm 0.001$ & $0.401 \pm 0.019$  & $0.365 \pm 0.014$ \\
EHRMamba & $0.300 \pm 0.007$ & $0.223 \pm 0.008$ & $0.912 \pm 0.014$ & $0.007 \pm 0.001$ & $0.391 \pm 0.009$  & $0.367 \pm 0.007$ \\
TALE-EHR & $0.353 \pm 0.016$ & $0.295 \pm 0.006$ & $\mathbf{0.960 \pm 0.002}$ & $0.006 \pm 0.001$ & $0.375 \pm 0.004$  & $0.333 \pm 0.010$ \\
\midrule
\textbf{MATA-Former} & $\mathbf{0.428 \pm 0.003}$ & $\mathbf{0.427 \pm 0.009}$ & $0.932 \pm 0.005$ & $\mathbf{0.005 \pm 0.001}$  & $\mathbf{0.496 \pm 0.011}$ & $\mathbf{0.456 \pm 0.010}$ \\
\bottomrule
\end{tabular}
\end{table*}

\begin{table*}[t]
\centering
\caption{Performance comparison on the MIMIC-IV v3.1 benchmark across three critical clinical prediction tasks. We report mean ± std over four independent runs. MATA-Former maintains robust stability and generalizability across different datasets}
\label{tab:MIMIC}
\small

\begin{tabular}{l ccccccccc}
\toprule

\multirow{2}{*}{\textbf{Model}} & \multicolumn{2}{c}{\textbf{Sepsis}} & \multicolumn{2}{c}{\textbf{IMV}} & \multicolumn{2}{c}{\textbf{Mortality}} \\
\cmidrule(lr){2-3} \cmidrule(lr){4-5} \cmidrule(lr){6-7} 
& AUPRC & AUROC & AUPRC & AUROC & AUPRC & AUROC \\
\midrule
HiTANet 
    & $ 0.563 \pm 0.018 $ & $0.861 \pm 0.009$ 
    & $0.557 \pm 0.031$ & $0.804 \pm 0.023$
    & $0.395 \pm 0.102$ & $0.864 \pm 0.036$  \\ 
EHRMamba 
    & $\mathbf{0.753 \pm 0.006 }$ & $\mathbf{0.948 \pm 0.020} $
    & $0.679 \pm 0.021$ & $0.830 \pm 0.021$ 
    & $0.385 \pm 0.036$ & $0.818 \pm 0.040$  \\
TALE-EHR
    & $0.657 \pm 0.065$ & $0.940 \pm 0.011$
    & $0.652 \pm 0.027$ & $0.920 \pm 0.025$
    & $0.501 \pm 0.111$ & $\mathbf{0.902 \pm 0.005}$
      \\
\midrule
\textbf{MATA-Former} 
    & $0.709 \pm 0.033$ & $ 0.929 \pm 0.016 $ 
    & $\mathbf{0.741 \pm 0.037}$ & $\mathbf{0.933 \pm 0.037}$ 
    & $\mathbf{0.523 \pm 0.091}$ & $0.899 \pm 0.003$ \\
\bottomrule
\end{tabular}%
\end{table*}

\subsection{Baselines}
To comprehensively evaluate MATA-Former, we conduct benchmarking on the SIICU dataset against methods ranging from foundational sequence models to state-of-the-art EHR architectures. We include Doctor AI~\cite{choi2016} and LSTM~\cite{lipton2015} as recurrent baselines, alongside a vanilla Transformer~\cite{shazeer2020} to control for time-aware mechanisms. Domain-specific comparisons include HiTANet~\cite{luo2020} for hierarchical time modeling, and EHRMamba~\cite{ehrmamba}, which adapts selective state space models for linear-complexity processing of long clinical trajectories. Finally, we benchmark against TALE-EHR~\cite{yu2025}, the current SOTA representing the frontier of the ``LLM Semantics + Time-Awareness'' paradigm. All data were processed identically as described in \cref{sec:unified_embedding}

\subsection{Performance Analysis}

\cref{tab:SIICU} presents the comparative performance on the SIICU dataset. Given the $<1\%$ positive prevalence, we prioritize AUPRC to address class imbalance. MATA-Former achieves a Sample AUPRC of $0.428$, significantly outperforming TALE-EHR ($0.353$) and the Transformer ($0.372$). A critical analysis of the results reveals a significant divergence between metric trends; baselines such as Doctor AI and LSTM attain near-perfect AUROC scores ($>0.930$) yet exhibit suboptimal AUPRC values ($<0.210$). This discrepancy indicates that recurrent architectures tend to overfit the dominant negative class while failing to capture the minority risk patterns essential for early warning systems. In contrast, MATA-Former's alignment mechanism effectively handles irregular intervals, yielding the lowest Brier Score ($0.005$) and a superior P@5 of $0.456$. This precision gap suggests that MATA-Former reduces the rate of false positives in high-confidence predictions, thereby addressing the practical challenge of clinical alarm fatigue.

\cref{tab:MIMIC} details robustness checks on MIMIC-IV. MATA-Former demonstrates strong generalizability, securing the highest AUPRC ($0.741$) and AUROC ($0.933$) in the IMV task, and outperforming baselines in Mortality prediction with an AUPRC of $0.523$. While EHRMamba shows competitive performance in Sepsis detection, its effectiveness fluctuates across different clinical targets. Conversely, MATA-Former maintains consistent stability across all three tasks, indicating that its architecture successfully captures complex semantic dependencies regardless of the underlying data distribution. This comparison suggests that baselines relying on static or strictly continuous time assumptions may struggle with domain shifts, whereas the proposed semantic-temporal alignment offers a more transferable solution for diverse clinical environments.  Experiment settings and stability analysis are provided in \cref{sec:hyperparameters} and \cref{sec:metric_stability}, respectively.

\subsection{Ablation Studies and Explanatory Analysis}
\label{sec:ablation_analysis}

\begin{figure} [t] 
    \centering
    \begin{subfigure}[b]{0.48\columnwidth} 
        \centering
        \includegraphics[width=\linewidth]{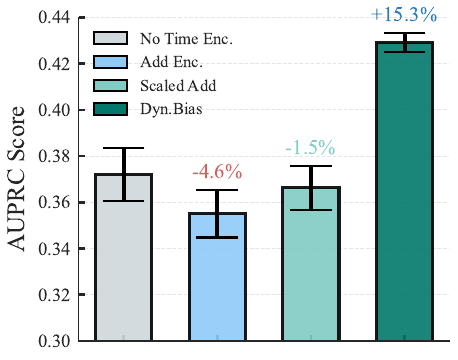}
        \caption{Temporal Encoding}
        \label{fig:3a}
    \end{subfigure}
    \hfill 
    \begin{subfigure}[b]{0.48\columnwidth} 
        \centering
        \includegraphics[width=\linewidth]{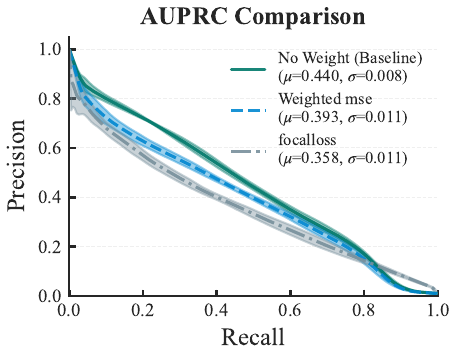}
        \caption{Loss Comparison}
        \label{fig:3b}
    \end{subfigure}
    \hfill 
    \begin{subfigure}[b]{0.48\columnwidth}
        \centering
        \includegraphics[width=\linewidth]{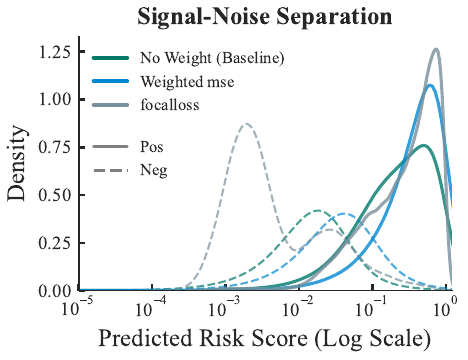}
        \caption{Signal-Noise Separation}
        \label{fig:3c}
    \end{subfigure}
    \hfill 
    \begin{subfigure}[b]{0.48\columnwidth}
        \centering
        \includegraphics[width=\linewidth]{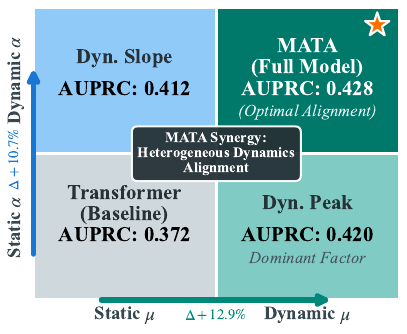}
        \caption{Geometric Ablation}
        \label{fig:3d}
    \end{subfigure}
    \caption{Ablation and mechanism analysis. (a) Additive bias superiorly preserves semantic integrity compared to temporal encoding. (b)-(c) MSE maximizes AUPRC by maintaining a distinct signal-noise separation valley.  (d) Decoupling $\mu$ and $\alpha$ achieves optimality in capturing heterogeneous clinical dynamics.}
    
\end{figure}

\textbf{Temporal Encoding Strategies.}  
Counter-intuitively, additive time encoding underperforms the time-agnostic baseline (TF) (\cref{fig:3a}). We attribute this degradation to two mechanisms: 
(1) Semantic Perturbation: Injecting time encodings directly disrupts the pre-trained semantic manifold—a distortion so severe that even substantial embedding scaling ($64\times$) fails to fully rectify it.
(2) Scale Invariance Violation: Absolute normalization ($t/T$) compromises physical consistency, mapping identical intervals to length-dependent values and obscuring causal latency(details in \cref{Time Encoding}). 
This also elucidates EHRMamba's underperformance on SIICU, where the aggressive injection of dense information into embedding representations disrupts the semantic manifold. In contrast, MATA decouples temporal injection via additive attention bias, preserving semantic integrity while aligning inductive bias with irregular clinical causality.

\begin{figure*}[t]
    \centering
    \begin{subfigure}{0.33\textwidth}
        \centering
        \includegraphics[width=\linewidth]{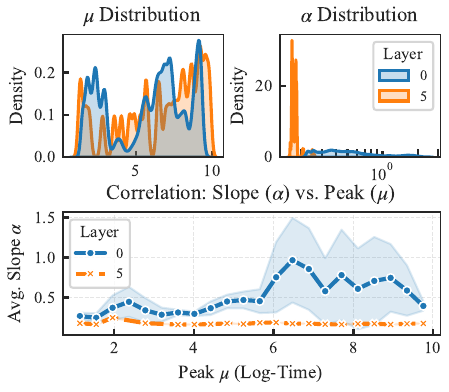}
        \caption{Dynamic Parameter Evolution}
        \label{fig:4a}
    \end{subfigure}
    \hfill 
    \begin{subfigure}{0.65\textwidth}
        \centering
        \includegraphics[width=\linewidth]{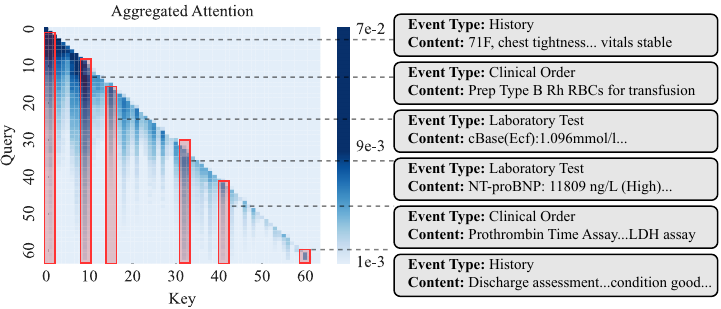}
        \caption{Aggregated Attention Alignment}
        \label{fig:4b}
    \end{subfigure}
    

    \caption{Visualization of learnable temporal dynamics. (a) Dynamic Parameter Evolution: The distributional shift of Laplacian parameters ($\mu, \alpha$) from Layer 0 to 5 reveals a transition from discrete temporal filtering to high-level semantic abstraction. (b) Aggregated Attention Alignment: To accommodate extended clinical trajectories, the heatmap utilizes $30 \times 30$ block-wise average pooling. The resulting vertical strata demonstrate consistent anchoring on pivotal prognostic determinants—specifically abnormal labs and interventional history—validating the structural alignment between the Laplacian prior and clinical causality.}
    \label{fig:attention_patterns_2_1_ratio}

\end{figure*}

\textbf{Loss Function.} In contrast to classification norms, re-weighting strategies (e.g., Focal Loss) prove counterproductive in our zero-inflated regression framework, yielding inferior AUPRC compared to Standard MSE. We attribute this to the over-penalization of hard negatives, thereby eroding signal-to-noise separation. As shown in \cref{fig:3b}, MSE sustains high precision in the top-tier regime (Recall $< 0.2$), whereas Focal Loss suffers early decay. \cref{fig:3c} reveals the mechanism: while MSE maintains a separation between noise and risk, Focal Loss induces a bimodal negative distribution. While it successfully concentrates easy negatives into a low-score peak, its aggressive focusing aggregates hard negatives into a high-score secondary peak that aliases with true signals. Consequently, our findings suggest that in extremely imbalanced regression, strict noise suppression proves more critical than aggressive positive boosting.

\textbf{Component Contribution.} \cref{fig:3d} validates the efficacy of dynamic parameterization. TF+Dyn.Slope (+10.7\%) confirms the utility of adaptive resolution, while the superior TF+Dyn.Peak (0.420 AUPRC) indicates that precise temporal localization yields higher information gain than bandwidth adjustment. We attribute the marginal gap between TF+Dyn.Peak and the full model to a geometric compensation effect: shifting the centroid $\mu$ on the logarithmic axis implicitly mimics decay rate $\alpha$ adjustments (proven in \cref{sec:parameter_interaction}). Ultimately, MATA-Former achieves optimality by decoupling these degrees of freedom, allowing independent control over search location ($\mu$) and focus precision ($\alpha$) to capture heterogeneous clinical dynamics.

\subsection{Interpretability and Visualization Analysis}
\label{sec:interpretability}

\textbf{Parameters Evolution.} To reveal the mechanism behind the model's emergent multi-scale temporal perception, we analyzed the layer-wise evolution of the geometric parameters $\mu_{i}^h$ and $\alpha_{i}^h$ (\cref{fig:4a}). The $\mu$ initially exhibits a continuous distribution across the logarithmic time axis, suggesting that the model captures a broad continuum of causal windows—ranging from immediate physiological reactions to sub-acute disease progression. Notably, in deeper layers (Layer 5), the $\mu$ distribution evolves into discrete modes, indicating that temporal anchors gradually lock onto specific time intervals. Concurrently, the $\alpha$ shifts significantly towards zero. This trend suggests a relaxation of local proximity constraints, effectively expanding the physical receptive field in deeper layers. 

To investigate the underlying mechanism more deeply, we projected the joint parameter distribution. The shallow layer (Layer 0) displays a diffuse, band-like pattern, indicative of a multi-resolution filter bank with rich temporal perception combinations. Here, we observe an adaptive bandwidth compensation mechanism: as the offset $\mu$ shifts towards distant history, $\alpha$ increases spontaneously. This positive coupling narrows the attention window to counteract the resolution dilution inherent to the logarithmic scale, preserving localization precision for remote events. Conversely, Layer 5 exhibits a structural collapse. The disappearance of the compensatory pattern marks a transition from physical calibration to semantic reasoning: by relaxing geometric distance penalties, deep layers transcend physical proximity, allowing attention to be driven primarily by medical causal logic for effective long-range integration.

\textbf{Clinical Alignment of Physical Receptive Fields.} In the initial layer (\cref{fig:4b}), structured activation patterns confirm the establishment of physical receptive fields. The distinct vertical continuity demonstrates that the network dynamically adjusts the offset $\mu$ to anchor attention on pivotal timestamps regardless of interval progression. This functions as a hard calibration mechanism, filtering noise through differentiated Laplacian horizons. Traceback analysis verifies that these high-attention strata autonomously align with high-entropy clinical events—specifically History, Clinical Order, Lab Test, and Radiology Report—rather than routine logs (details in  \cref{case ana}). This alignment validates the MATA mechanism, confirming the model's capacity to identify and sustain focus on pivotal clinical events via dynamic temporal modulation. Beyond quantitative precision, MATA-Former bridges the gap between deep learning and clinical reasoning through inherent transparency. The explicit parameterization of the Laplacian prior transforms attention mechanisms into interpretable temporal horizons, mimicking how physicians weigh acute fluctuations against chronic history during differential diagnosis. By mapping the learned decay rates $\alpha$ and offset $\mu$ to physical time, clinicians can verify that risk predictions stem from logically relevant temporal windows rather than spurious correlations. This structural alignment ensures that the model's decision boundaries respect pathological causality, fostering the trust required for deployment in high-stakes ICU environments.

\section{Limitations}
\label{sec:limitations}

Despite the superior performance of MATA-Former, this study remains subject to several limitations. 
(1) Single-Center Data Bias: Although we utilized a high-fidelity dataset, the data originates exclusively from a single institution. The distinctive charting patterns of this specific medical environment may compromise the model's generalization robustness when facing cross-center distributional shifts. 
(2) Noise Interference in Interpretability: While the MATA mechanism provides physical calibration, the model occasionally focuses on non-causal noise or spurious correlations. This indicates that there is still room for improvement in distinguishing between true ``pathological anchors'' and data artifacts. 
(3) Fragility of the Semantic Manifold: While Qwen3 embeddings effectively compress sparse, long-sequence clinical data, the resulting latent space exhibits inherent fragility. Direct superposition of unprocessed vectors without rigorous calibration disrupts the semantic manifold, leading to suboptimal convergence. This instability imposes strict constraints on the feasibility of performing arithmetic operations directly within the event vector space.
(4) Insufficient Modal Completeness: The unified textualization framework has not yet integrated high-frequency physiological waveforms (e.g., ECG) or medical imaging. This singular reliance on the textual modality constitutes an information bottleneck, potentially leading to the omission of certain latent pathological dynamics.

\section{Conclusions}
\label{sec:conclusion}

This paper proposes MATA-Former to bridge the semantic-temporal gap in ICU risk forecasting. By parameterizing pathological latencies, the MATA mechanism harmonizes unstructured clinical semantics with physical time dynamics. Complementing this, we introduce PSL, which utilizes Gaussian mapping to shift supervision from coarse binary classification to continuous regression, thereby capturing granular risk trajectories. To address data scarcity, we constructed SIICU, a high-fidelity dataset comprising over 506k expert-verified events. Evaluations confirm that MATA-Former delivers improvements- on the SIICU dataset while maintaining robust consistency across MIMIC-IV benchmarks, thereby validating the architecture's generalization capabilities beyond specific clinical settings. Future initiatives will focus on extending modal completeness to enhance the robustness and interpretability of next-generation Clinical Decision Support Systems.

\section*{Impact Statement}

This paper presents a deep learning framework dedicated to risk prediction in the Intensive Care Unit (ICU). The primary objective is to assist clinicians by providing timely and interpretable early warnings, thereby potentially reducing mortality rates and mitigating information overload. However, we acknowledge several critical ethical considerations. 

First, despite its superior performance, the model is designed strictly as a Clinical Decision Support System and must never substitute for human judgment. False negatives may delay necessary interventions, while false positives contribute to alarm fatigue; therefore, deployment necessitates rigorous validation and the implementation of ``human-in-the-loop'' protocols. 

Second, as the model was trained on single-center data, it may harbor demographic or institutional biases. Cross-center validation is indispensable to ensure fairness prior to widespread application. 

Finally, while all patient data utilized in this study underwent rigorous de-identification and received approval from the Medical Ethics Committee (Approval No.[2025YS-035]), future applications must strictly adhere to data privacy regulations and informed consent protocols.

\bibliographystyle{icml2026}
\bibliography{icml2026/reference}


\newpage
\appendix
\onecolumn 
\crefalias{section}{appendix}
\crefalias{subsection}{appendix}

\section{Mathematical Foundations and Empirical Analysis of MATA Mechanism}
\label{sec:appendix_a}

\label{sec:sensitivity_analysis}

\subsection{Theoretical Bounds on Effective Attention Horizon}

To quantify the Effective Attention Horizon—the temporal boundary beyond which historical context becomes numerically negligible—we analyze the decay characteristics of the dynamic Laplacian bias $\mathcal{B}_{ij}^{(h)}$.

The attention weight $A_{ij}^{(h)}$ for the $h$-th head is defined under the additive formulation as:
\begin{equation}
    A_{ij}^{(h)} = \frac{\exp \left( \mathcal{S}_{ij}^{(h)} + \mathcal{B}_{ij}^{(h)} + \mathcal{M}_{ij}^{\text{Causal}} \right)}{\sum_{l} \exp \left( \mathcal{S}_{il}^{(h)} + \mathcal{B}_{il}^{(h)} + \mathcal{M}_{il}^{\text{Causal}} \right)},
\end{equation}
where $\mathcal{S}_{ij}^{(h)} = \mathbf{q}_i^{(h)} (\mathbf{k}_j^{(h)})^\top / \sqrt{d_k}$ represents the semantic affinity.

To isolate the influence of the Laplacian prior, we examine a baseline scenario with uniform semantic affinity, assuming $\mathcal{S}_{ij}^{(h)} = C$ for all causal steps ($t_j \le t_i$). Let $j^*$ denote the optimal temporal anchor where the normalized log-distance aligns with the peak offset ($D_{ij^*} = \mu_{i}^{(h)}$), resulting in a zero bias penalty ($\mathcal{B}_{ij^*}^{(h)} = 0$).

We introduce the \textit{Relative Attention Ratio} $R(\Gamma)$, defined as the attention mass at an arbitrary decay point $j$ normalized by the peak mass at anchor $j^*$. With the penalty magnitude given by $\Gamma = \alpha_{i}^{(h)} |D_{ij} - \mu_{i}^{(h)}|$, the ratio simplifies to:
\begin{equation}
    R(\Gamma) \triangleq \frac{A_{ij}^{(h)}}{A_{ij^*}^{(h)}} = \frac{\exp(C - \Gamma)}{\exp(C)} = \exp(-\Gamma).
\end{equation}

This relationship maps the penalty magnitude directly to attention suppression. We establish the effective horizon using two numerical significance thresholds:

\begin{enumerate}
    \item \textit{1\% retention threshold} ($R = 10^{-2}$): The penalty required to attenuate the attention weight to 1\% of its peak is calculated as:
    \begin{equation}
        \Gamma_{1\%} = -\ln(0.01) \approx 4.605.
    \end{equation}
    \item \textit{0.1\% retention threshold} ($R = 10^{-3}$): Adopting a stricter standard often used as a proxy for numerical zero, the required penalty is:
    \begin{equation}
        \Gamma_{0.1\%} = -\ln(0.001) \approx 6.908.
    \end{equation}
\end{enumerate}

Guided by the sensitivity thresholds, we establish $\Gamma = 5$ as the operational cutoff for the effective attention horizon. At this level, the relative attention ratio decays to $R(5) \approx 6.7 \times 10^{-3}$, indicating that over $99.3\%$ of the peak attention mass is suppressed. Consequently, for the $h$-th head, a historical position $j$ lies beyond the effective temporal receptive field of query $\mathbf{q}_i^{(h)}$ when the normalized log-distance satisfies:
\begin{equation}
\alpha_{i}^{(h)} \cdot |D_{ij} - \mu_{i}^{(h)}| > 5
\end{equation}

\subsection{Inverse Mapping to Physical Time}
\label{sec:physical_mapping}

To map the learned temporal dynamics back to clinically interpretable units, we project the boundaries from the logarithmic domain onto the physical time axis. This inversion quantifies the duration spanned by the temporal bias $\mathcal{B}_{ij}^{(h)}$, facilitating a comparison with established clinical validity windows.

Given the cutoff threshold $\Gamma$, the effective attention window in logarithmic space is constrained by the inequality $\alpha_{i}^{(h)} |D_{ij} - \mu_{i}^{(h)}| \le \Gamma$. With a strictly positive decay rate $\alpha_{i}^{(h)}$, we define the logarithmic effective radius $X_{i}^{(h)}$ as the half-width of the receptive field around the peak $\mu_{i}^{(h)}$:
\begin{equation}
    X_{i}^{(h)} = \frac{\Gamma}{\alpha_{i}^{(h)}}.
\end{equation}
This establishes the logarithmic interval $D_{ij} \in [\mu_{i}^{(h)} - X_{i}^{(h)}, \mu_{i}^{(h)} + X_{i}^{(h)}]$. To recover the physical time span, we invert the normalized log-distance function $D_{ij} = \ln(\Delta t / \tau + 1)$, where $\Delta t = |t_i - t_j|$ represents the time lag and $\tau = 60$ serves as the scaling factor.

The proximal attention boundary $T_{\min}$ represents the minimum temporal lag required for significant attention activation, effectively delineating the near-term exclusion zone. By equating the normalized distance to the lower logarithmic limit $D_{ij} = \mu_{i}^{(h)} - X_{i}^{(h)}$ and solving for the time difference, we obtain:
\begin{equation}
    T_{\min} = \max \left( 0, \, \tau \left[ \exp \left( \mu_{i}^{(h)} - X_{i}^{(h)} \right) - 1 \right] \right).
\end{equation}
The rectification operation ensures the boundary respects physical causality, as the model cannot attend to future events relative to the query.

Conversely, the distal attention boundary $T_{\max}$ defines the maximum historical horizon, setting the limit for information retrieval. This boundary corresponds to the upper logarithmic limit $D_{ij} = \mu_{i}^{(h)} + X_{i}^{(h)}$:
\begin{equation}
    T_{\max} = \tau \left[ \exp \left( \mu_{i}^{(h)} + X_{i}^{(h)} \right) - 1 \right].
\end{equation}

This inverse transformation highlights a critical structural property: while the Laplacian bias profile remains symmetric around $\mu_{i}^{(h)}$ within the logarithmic manifold, its projection onto the physical time axis manifests as a highly asymmetric, heavy-tailed distribution. Due to the convexity of the mapping function $f(D) = \tau(e^D - 1)$, equal displacements $\pm X$ on the logarithmic axis produce distinct effects on the physical axis. Steps towards the past ($\mu + X$) are exponentially amplified in physical time, while steps towards the present ($\mu - X$) are severely compressed. This allows the model to automatically acquire a broader temporal tolerance when attending to distant history.

Table \ref{tab:receptive_field_examples} illustrates the physical receptive fields corresponding to different parameter combinations under the condition $\Gamma = 5, \tau = 60$.

\begin{table}[h]
\centering
\caption{Examples of Physical Receptive Fields under different parameter combinations ($\Gamma = 5, \tau = 60$). These values demonstrate how the interaction of $\mu$ and $\alpha$ maps to clinically meaningful time windows.}
\label{tab:receptive_field_examples}
\setlength{\tabcolsep}{4pt}
\begin{tabular}{@{}cccccc@{}}
\toprule
\textbf{Peak ($\mu$)} & \textbf{Slope ($\alpha$)} & \textbf{Log Radius ($X$)} & \textbf{Physical Range $[T_{min} ,T_{max}]$}  & \textbf{Clinical Semantic Interpretation} \\ \midrule
1.0 & 2 & 2.5 & 0 sec -- 32.2 min  & Captures minute-level fluctuations \\
5.0 & 2 & 2.5 & 11.2 min -- 30.1 hours  & Captures diurnal trends \\
9.5 & 2 & 2.5 & 18.2 hours -- 113.1 days  &  Captures seasonal patterns \\ \bottomrule
\end{tabular}
\end{table}

\begin{figure*}[t!] 
    \centering
    
    \begin{subfigure}[t]{0.32\textwidth}
      \centering
      \includegraphics[width=\linewidth]{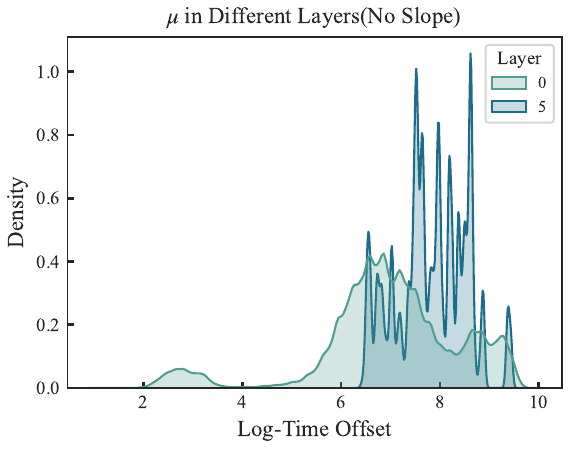}
      \caption{No-Slope: $\mu$ Distribution}
      \label{fig:noslope_intervals} 
    \end{subfigure}
    \hfill
    \begin{subfigure}[t]{0.32\textwidth}
      \centering
      \includegraphics[width=\linewidth]{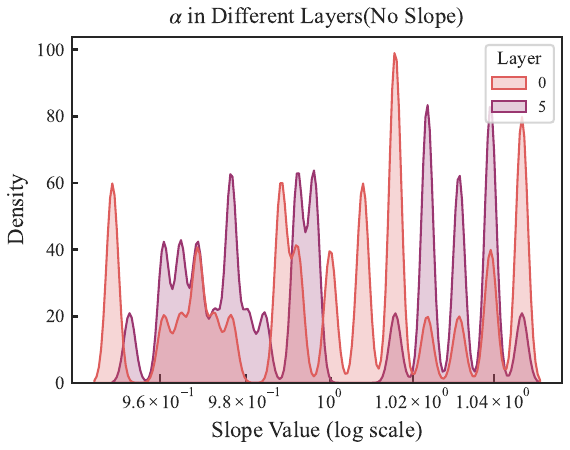}
      \caption{No-Slope: $\alpha$ Distribution}
      \label{fig:noslope_los}
    \end{subfigure}
    \hfill
    \begin{subfigure}[t]{0.32\textwidth}
      \centering
      \includegraphics[width=\linewidth]{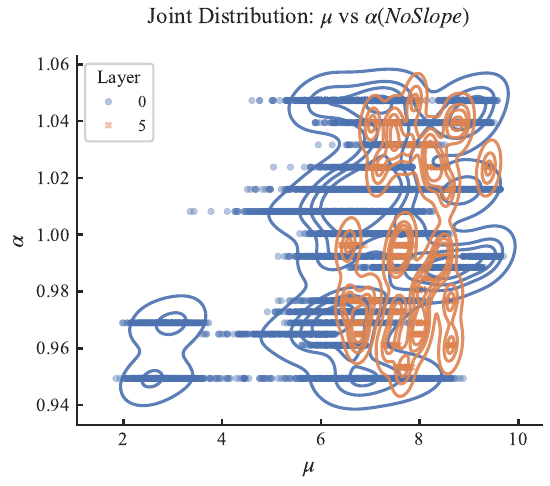}
      \caption{No-Slope: Joint Distribution}
      \label{fig:noslope_params}
    \end{subfigure}
    
    \vspace{0.1in} 
    
    \begin{subfigure}[t]{0.32\textwidth}
      \centering
      \includegraphics[width=\linewidth]{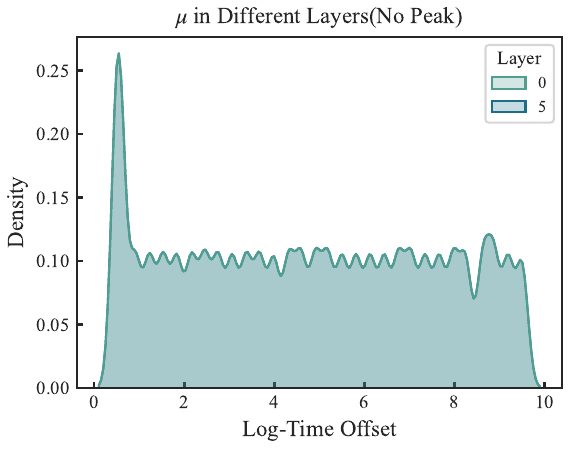}
      \caption{No-Peak: $\mu$ Distribution}
      \label{fig:nopeak_intervals}
    \end{subfigure}
    \hfill
    \begin{subfigure}[t]{0.32\textwidth}
      \centering
      \includegraphics[width=\linewidth]{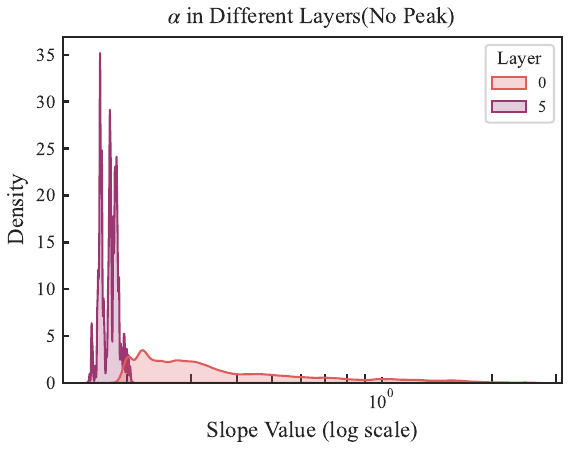}
      \caption{No-Peak: $\alpha$ Distribution}
      \label{fig:nopeak_los}
    \end{subfigure}
    \hfill
    \begin{subfigure}[t]{0.32\textwidth}
      \centering
      \includegraphics[width=\linewidth]{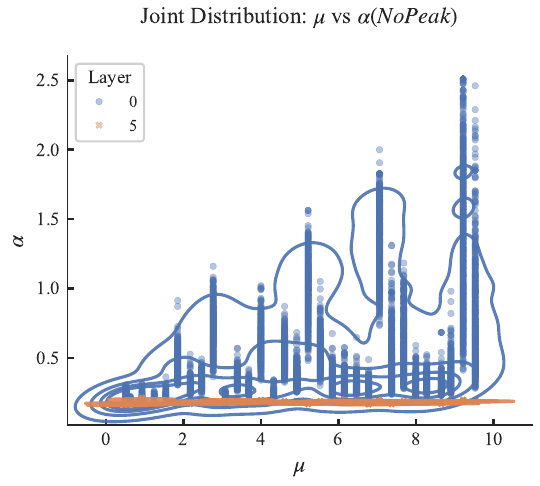}
      \caption{No-Peak: Joint Distribution}
      \label{fig:nopeak_params}
    \end{subfigure}

    \vspace{0.1in}

    \begin{subfigure}[t]{0.32\textwidth}
      \centering
      \includegraphics[width=\linewidth]{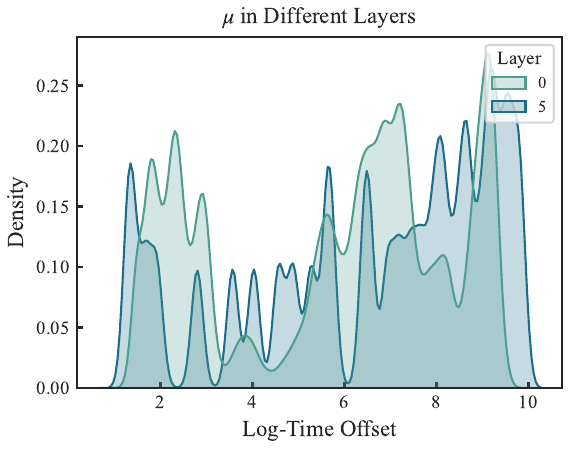}
      \caption{Full: $\mu$ Distribution}
      \label{fig:all_intervals}
    \end{subfigure}
    \hfill
    \begin{subfigure}[t]{0.32\textwidth}
      \centering
      \includegraphics[width=\linewidth]{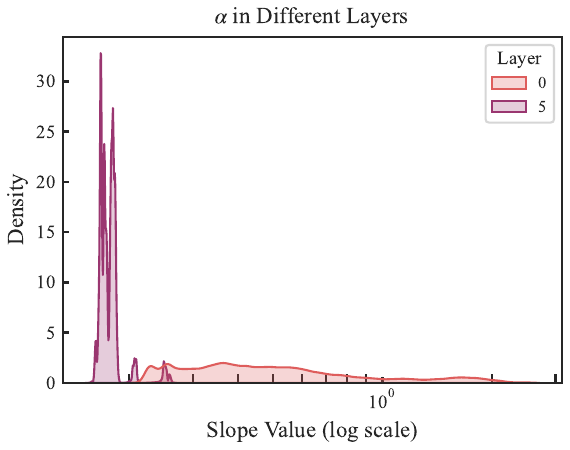}
      \caption{Full: $\alpha$ Distribution}
      \label{fig:all_los}
    \end{subfigure}
    \hfill
    \begin{subfigure}[t]{0.32\textwidth}
      \centering
      \includegraphics[width=\linewidth]{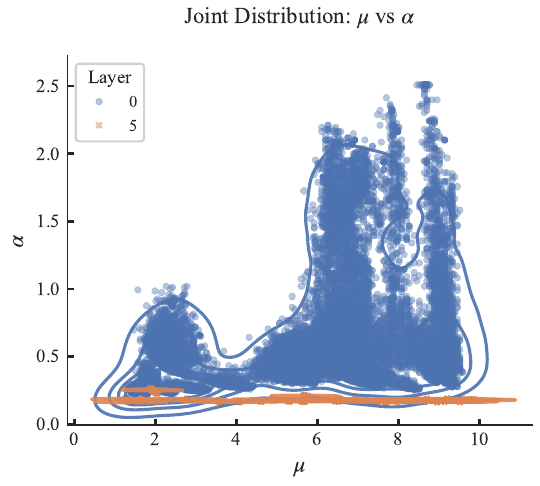}
      \caption{Full: Joint Distribution}
      \label{fig:all_params}
    \end{subfigure}
    
    \caption{Visualization of bias parameters. (Top) No-Slope setting; (Middle) No-Peak setting; (Bottom) Full model configuration. The columns represent $\mu$ Distribution, $\alpha$ Distribution, and Joint Distribution respectively.}
    \label{fig:attention_patterns}
\end{figure*}

\subsection{Compensatory Effect between Peak (\texorpdfstring{$\mu$}{mu}) and Slope (\texorpdfstring{$\alpha$}{alpha})}
\label{sec:parameter_interaction}

Within the MATA mechanism, the parameters $\mu_{i}^{(h)}$ and $\alpha_{i}^{(h)}$ jointly define the morphology of the temporal bias. However, ablation studies suggest that fixing one of these parameters results in minimal performance degradation. We elucidate this redundancy by deriving the functional overlap and compensatory mechanism inherent to the physical time scale.

Unlike traditional linear temporal models where positional shifting and scaling are decoupled, the logarithmic projection $D = \ln(\Delta t / \tau + 1)$ introduces a nonlinear coupling between the peak location and the receptive width. Building on the physical boundary formulations from \cref{sec:physical_mapping}, we define the physical attention bandwidth $BW$ as the difference between the distal and proximal limits:
\begin{equation}
    BW = T_{\max} - T_{\min} = \tau e^{\mu_{i}^{(h)}} \left( e^{X_{i}^{(h)}} - e^{-X_{i}^{(h)}} \right).
\end{equation}
Substituting the effective radius $X_{i}^{(h)} = \Gamma / \alpha_{i}^{(h)}$, we obtain the complete analytical expression for the bandwidth:
\begin{equation}
    BW = \tau e^{\mu_{i}^{(h)}} \left[ \exp \left( \frac{\Gamma}{\alpha_{i}^{(h)}} \right) - \exp \left( - \frac{\Gamma}{\alpha_{i}^{(h)}} \right) \right].
\end{equation}

The bandwidth expression reveals a dual sensitivity to the controlling parameters:
\begin{itemize}
    \item \textit{Scaling mechanism:} Decreasing the decay rate $\alpha_{i}^{(h)}$ amplifies the bracketed magnitude, thereby expanding the bandwidth.
    \item \textit{Translation mechanism:} Increasing the offset $\mu_{i}^{(h)}$ scales the physical range exponentially via the factor $e^{\mu_{i}^{(h)}}$.
\end{itemize}

Crucially, a translational shift of $\mu_{i}^{(h)}$ in the logarithmic domain manifests as a multiplicative scaling effect in the physical domain, inducing a partial functional overlap within the parameter space. Consequently, constraining the slope to a constant (e.g., $\alpha \approx 1.0$) does not compromise the model's capacity to regulate the physical field of view. With $\alpha_{i}^{(h)}$ fixed, the sensitivity of the bandwidth to the offset satisfies:
\begin{equation}
    \frac{\partial BW}{\partial \mu_{i}^{(h)}} = BW.
\end{equation}
This relationship confirms that the physical bandwidth grows exponentially with $\mu_{i}^{(h)}$. To capture long-range dependencies, the model need not adjust the decay rate; simply shifting the temporal center $\mu_{i}^{(h)}$ deeper into history provides sufficient bandwidth expansion. Due to the exponential nature of the term $e^{\mu_{i}^{(h)}}$, this compensatory shift effectively covers diverse clinical scales, ranging from hours to months, preserving the expressive power of the fixed-slope configuration.

Experimental findings substantiate these mathematical derivations. In the static peak configuration ($\bar{\mu}$) with a dynamic slope, the parameter $\alpha$ in deeper network layers exhibits a pronounced collapse toward minimal values ($\alpha \to 0.2$). This evolutionary trajectory closely parallels the behavior of the full model. Constrained by a fixed $\mu$, the mechanism relaxes physical proximity constraints by compressing $\alpha$, thereby extending the effective receptive field around static temporal anchors. This morphological consistency confirms that deep layers prioritize global semantic integration; even without translational flexibility, the model achieves cross-order temporal coverage purely through resolution modulation.

Conversely, under the static slope configuration ($\bar{\alpha}$), the system secures sufficient bandwidth compensation via the growth of the offset $\mu$. Leveraging the exponential scaling effect established in the bandwidth derivation, the model successfully accommodates observational horizons spanning from minutes to months. This compensatory capability elucidates why the fixed-slope variant retains robust expressive power despite its parametric restrictions.

\section{Experimental Details}

\subsection{Parameter Initialization and Search Space Design}
\label{mu_alpha_Initialization}
To ensure the MCE-Aware attention mechanism effectively captures multi-scale clinical features while maintaining numerical stability, we designed a rigorous parameter generation pipeline that maps the architectural logic directly to the semantic space. The predictor $\mathcal{F}_\phi$ processes the query vector via a two-layer Multi-Layer Perceptron (hidden dimension 64, Tanh activation) to output residual components $\delta_{\alpha}$ and $\delta_{\mu}$. To enforce a warm start governed purely by the static prior, the weights and biases of the final projection layer are initialized to zero. The geometric parameters are subsequently generated through modulation mechanisms designed to align with clinical semantics.

\paragraph{Slope Parameter ($\alpha$).}
For the decay rate $\alpha_{i}^{(h)}$, we employ a multiplicative modulation scheme to achieve cross-order resolution scaling. To ensure numerical stability, the output is bounded by hard constraints:
\begin{equation}
    \alpha_{i}^{(h)} = \text{Clamp}\left(\bar{\alpha}^{(h)} \cdot \exp(\delta_{\alpha, i}^{(h)}), \epsilon, \gamma_{\alpha}^{\text{limit}}\right).
\end{equation}
Consistent with the implementation, we set the lower bound $\epsilon = 10^{-4}$ and the upper limit $\gamma_{\alpha}^{\text{limit}} = 2.5$. We initialize the priors $\bar{\alpha}^{(h)}$ with uniform perturbations within $[-0.05, 0.05]$ around a baseline of $1.0$. The exponential mapping ensures symmetric gradient optimization for field expansion and contraction, allowing the model to transition dynamically between high-precision focusing ($\alpha \to 2.5$) and global trend integration ($\alpha \to 0$).

\paragraph{Positional Parameter ($\mu$).}
For the center shift $\mu_{i}^{(h)}$, we implement a translation of the focal point via residual updates in the logit space:
\begin{equation}
    \mu_{i}^{(h)} = \sigma\left(\text{logit}(\bar{\mu}^{(h)} / \gamma_{\mu}) + \lambda \cdot \delta_{\mu, i}^{(h)}\right) \cdot \gamma_{\mu}.
\end{equation}
A sensitivity coefficient $\lambda = 4$ is introduced to amplify the influence of the residual component, enabling the attention focus to rapidly lock onto core time windows. We set the upper bound $\gamma_{\mu} = 10.0$, which maps to a physical history of approximately $15.3$ days (given $\tau=60$). To ensure structured coverage of this temporal domain, the priors $\bar{\mu}^{(h)}$ are linearly tiled across the interval $[0, 0.95 \gamma_{\mu}]$. Note that during initialization, these probability values are internally clamped to $[0.05, 0.95]$ before logit transformation to prevent numerical overflow, ensuring robust starting points for optimization.


\begin{figure}[p] 
    \centering
    
    \captionof{table}{Representative sampling points of $\mu_{i}^{(h)}$ within the search space and their corresponding clinical interpretations.}
    \label{tab:mu_sampling_points} 
    \small
    \setlength{\tabcolsep}{4pt}
    \begin{tabular}{@{}cccc@{}}
    \toprule
    \textbf{Peak ($\mu_{i}^{(h)}$)} & \textbf{Physical Time ($\Delta t$)} & \textbf{Clinical Unit} & \textbf{Semantics} \\ \midrule
    0 & 0 s & Real-time & Current vital signs \\
    4.1 & $\approx$ 3,600 s & 1 Hour & Acute evolution \\
    7.1 & $\approx$ 72,000 s & 20 Hours & Diurnal cycle \\
    10 & $\approx$ 1,321,500 s & 15.3 Days & Long-term history \\ \bottomrule
    \end{tabular}
    
    \vspace{0.6cm} 

    \captionof{table}{Detailed Hyperparameter Configuration for MATA-Former.}
    \label{tab:hyperparameters}
    \small
    \begin{tabular}{ll}
    \toprule
    \textbf{Category} & \textbf{Specification} \\
    \midrule
    \multicolumn{2}{l}{\textit{Software Environment}} \\
    Operating System & Linux \\
    Python Version & 3.12.11 \\
    PyTorch Version & 2.8.0 \\
    Accelerate Version & 1.11.0 \\
    DeepSpeed Version & 0.18.1 \\
    Transformers Version & 4.57.1 \\
    \midrule
    \multicolumn{2}{l}{\textit{Distributed Training}} \\
    Distributed Strategy & DeepSpeed ZeRO Stage-2 \\
    Number of Processes & 8 \\
    Mixed Precision & BFloat16 \\
    Gradient Checkpointing & Enabled \\
    \midrule
    \multicolumn{2}{l}{\textit{Model Architecture}} \\
    Model Type & MATA-Former \\
    Hidden Dimension ($d_{model}$) & 4,096 \\
    Number of Layers & 6 \\
    Attention Heads & 32 \\
    Feed-forward Dimension & 11,008 \\
    Output Dimension & 1,440 \\
    Normalization & RMSNorm \\
    Activation Function & SiLU \\
    \midrule
    \multicolumn{2}{l}{\textit{Training Hyperparameters}} \\
    Optimizer & AdamW \\
    Base Learning Rate & $5 \times 10^{-5}$ \\
    Predictor Learning Rate & $5 \times 10^{-4}$ \\
    Weight Decay & 0.01 \\
    LR Scheduler & Cosine with Warmup \\
    Warmup Ratio & 5\% \\
    Global Batch Size & 8 \\
    Maximum Epochs & 750 \\
    Early Stopping Patience & 50 epochs \\
    Cross-Validation & 4-Fold \\
    \bottomrule
    \end{tabular}
    
    \vspace{0.6cm}

    
    \includegraphics[width=0.40\linewidth]{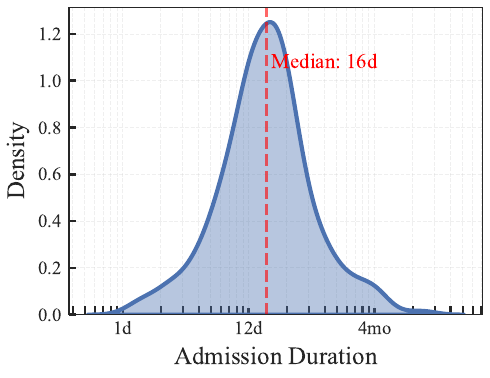}
    \captionof{figure}{Statistics on the length of time patients stay in the ICU.}
    \label{fig:6a} 
    \vspace{-12pt}
\end{figure}

\subsection{Experimental Environment and Hardware Configuration}
\label{sec:hardware_config}

To ensure the reproducibility of our experiments, we detailed the computational environment. All experiments were conducted on a high-performance computing node equipped with 8 NVIDIA H100 (80GB) GPUs. The software environment was based on PyTorch 2.x, utilizing the Hugging Face Accelerate framework in conjunction with DeepSpeed ZeRO Stage 2 optimization techniques for distributed training. By employing \textit{Gradient Checkpointing} technology, we effectively mitigated the GPU memory pressure caused by long-sequence clinical time-series data while maintaining a high-dimensional representation of $d_{model}=4096$.

\subsection{Hyperparameter Settings, Training Strategy, and Optimization}
\label{sec:hyperparameters}

The MATA-Former utilizes a deep Transformer architecture distinguished by its adaptive temporal perception. The network comprises 6 Transformer blocks, each incorporating 32 attention heads. To balance parameter efficiency with representational capacity, we establish a hidden dimension of $d_{model} = 4096$ and a Feed-Forward Network intermediate dimension of 11,008.

To accommodate the distinct optimization dynamics of different modules, we employ a heterogeneous parameter update strategy using the AdamW optimizer with a weight decay of 0.01. We assign a differential learning rate to the parameter prediction network $\mathcal{F}_\phi$. Since $\mathcal{F}_\phi$ is lightweight yet governs sensitive temporal dynamics via additive peak translations, a standard learning rate could lead to negligible updates relative to the massive backbone gradients. Consequently, we set the learning rate for $\mathcal{F}_\phi$ to $5 \times 10^{-4}$, ten times that of the backbone ($5 \times 10^{-5}$). This configuration accelerates convergence toward the optimal temporal bias morphology while preserving the stability of the underlying semantic features.

We implement a Cosine Annealing schedule supplemented by a 5\% warmup period. This approach stabilizes the initial exploration of the parameter space, effectively mitigating gradient oscillations potentially arising from the initialization of the MATA bias terms. Comprehensive hyperparameter configurations are provided in Table \ref{tab:hyperparameters}.

\begin{algorithm}[tb]
   \caption{MCE-Aware Time-ALiBi Attention}
   \label{alg:mce_attention}
\begin{algorithmic}[1] 
   \STATE {\bfseries Input:} Hidden states $\mathbf{H} \in \mathbb{R}^{B \times S \times D}$, timestamps $\mathbf{T} \in \mathbb{R}^{B \times S}$
   \STATE {\bfseries Input:} Per-head priors: slope $\boldsymbol{\alpha}^{(0)} \in \mathbb{R}^{H}$, peak $\boldsymbol{\mu}^{(0)} \in \mathbb{R}^{H}$
   \STATE {\bfseries Output:} Output $\mathbf{O} \in \mathbb{R}^{B \times S \times D}$
   \STATE
   \STATE \textbf{// Standard multi-head QKV projection}
   \STATE $\mathbf{Q}, \mathbf{K}, \mathbf{V} \gets \text{Linear}(\mathbf{H})$ \COMMENT{Reshape to $\mathbb{R}^{B \times H \times S \times d_h}$}
   \STATE $\mathbf{E} \gets \mathbf{Q}\mathbf{K}^\top / \sqrt{d_h}$ \COMMENT{Content-based attention}
   \STATE
   \STATE \textbf{// Log-scaled temporal distance matrix}
   \STATE $\mathbf{D} \gets \log(|\mathbf{T}_i - \mathbf{T}_j| / 60 + 1)$
   \STATE
   \STATE \textbf{// Key Innovation: Query-adaptive temporal parameters}
   \STATE $[\Delta\alpha, \Delta\mu] \gets f_{\text{MCE}}(\mathbf{Q})$ \COMMENT{MLP predicts per-query adjustments}
   \STATE $\boldsymbol{\alpha} \gets \boldsymbol{\alpha}^{(0)} \cdot \exp(\Delta\alpha)$ \COMMENT{Dynamic slope}
   \STATE $\boldsymbol{\mu} \gets \sigma(\boldsymbol{\mu}^{(0)} + \Delta\mu) \cdot \mu_{\max}$ \COMMENT{Dynamic peak}
   \STATE
   \STATE \textbf{// temporal attention bias}
   \STATE $\mathbf{B}_{\text{time}} \gets -\boldsymbol{\alpha} \cdot |\mathbf{D} - \boldsymbol{\mu}|$
   \STATE
   \STATE \textbf{// Combine and compute attention}
   \STATE $\mathbf{P} \gets \text{Softmax}(\mathbf{E} + \mathbf{B}_{\text{time}} + \mathbf{M}_{\text{causal}})$
   \STATE $\mathbf{O} \gets \text{Linear}(\text{Concat}(\mathbf{P}\mathbf{V}))$
   \STATE \textbf{return} $\mathbf{O}$
\end{algorithmic}
\end{algorithm}

\begin{figure*}
    \centering
    \includegraphics[width=\linewidth]{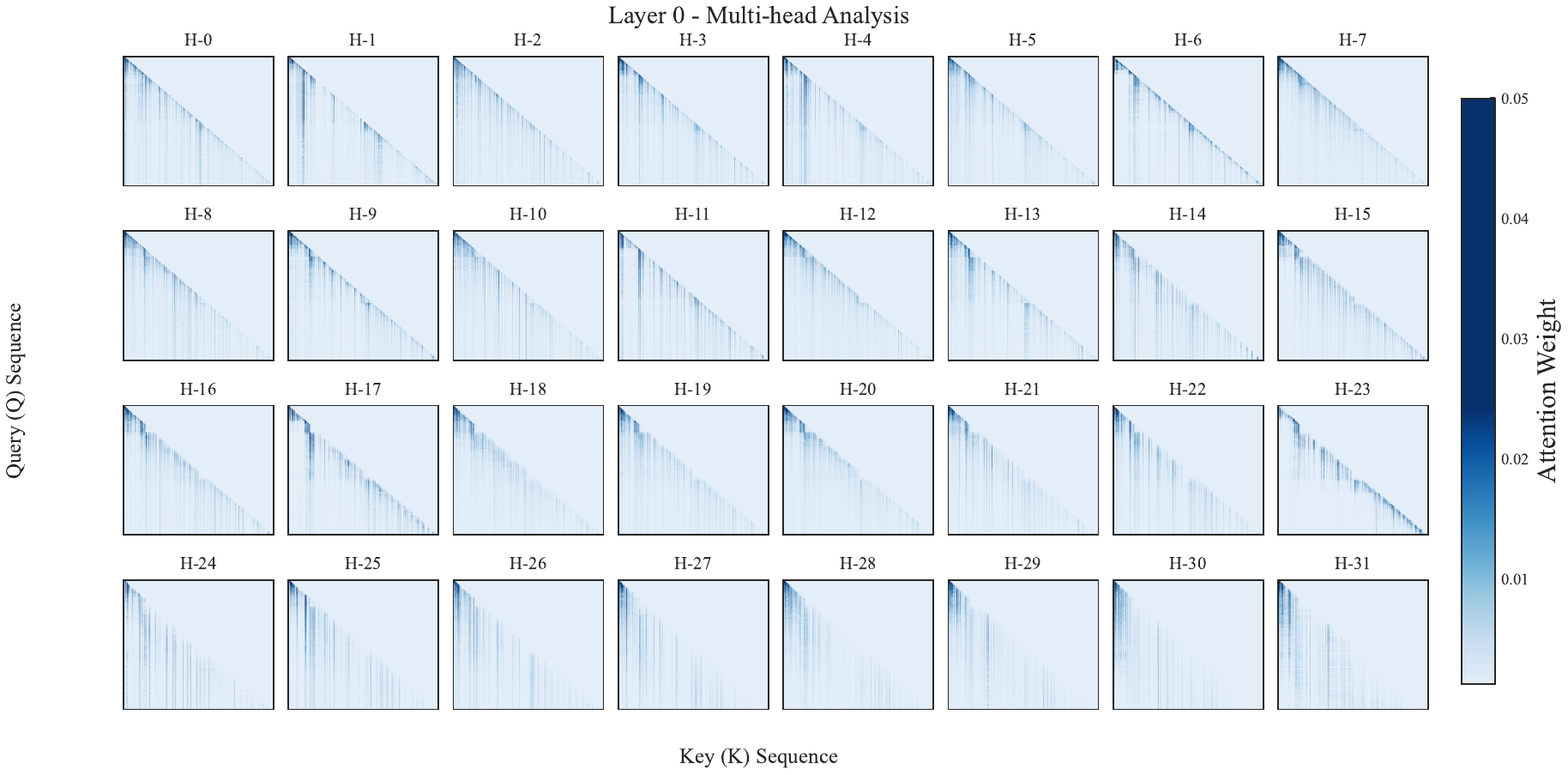}
    \caption{Visualization of Multi-Head Attention Patterns in Layer 0. The figure displays the attention weight matrices for all 32 heads. The heatmaps illustrate the strictly causal structure (lower triangular mask) and the diversity of attention distributions, where varying degrees of concentration indicate different effective receptive fields learned by the MATA mechanism.}
    \vspace{-1.0\baselineskip}
\end{figure*}

\section{Rationality Analysis and Case Verification}

\subsection{Multi-Head Aggregation as Mixture of Laplacians}

A theoretical concern regarding the proposed MATA mechanism implies that the inherent unimodality of the Laplacian prior $\mathcal{B}_{ij}^{(h)}$ might constrain a single attention head to a contiguous temporal segment centered at $\mu_{i}^{(h)}$. However, complex clinical reasoning frequently necessitates the simultaneous correlation of disjoint temporal windows, such as linking admission baselines to acute reactions occurring days later. We demonstrate that MATA-Former circumvents this limitation through multi-head aggregation, which effectively functions as a Mixture of Laplacians.

Formally, the final context representation $\mathbf{o}_i$ is derived via the linear projection of concatenated head outputs: $\mathbf{o}_i = \sum_{h=1}^{H} \mathbf{W}_h^O \text{Attention}_h(\mathbf{Q}, \mathbf{K}, \mathbf{V})$. Since each head $h$ independently regresses a distinct offset $\mu_{i}^{(h)}$ and decay rate $\alpha_{i}^{(h)}$, the aggregate temporal receptive field approximates a multi-modal probability density function. As evidenced in the layer-wise visualization, different heads exhibit functional orthogonality; for instance, ``Immediate Reflex Monitors'' (e.g., Head-0, $\mu \approx 3$ min) and ``Long-term Historians'' (e.g., Head-24, $\mu \approx 2$ days) activate concurrently for the same query. By superimposing these distinct unimodal distributions, the model reconstructs a complex, non-contiguous temporal dependency map, confirming that the multi-head ensemble provides a universal approximation capability for irregular clinical trajectories.

\begin{small} 
\begin{longtable}{lp{2cm}p{10cm}} 
    
    \caption{Detailed Events and Medical History Records.} 
    \label{tab:Key_events} \\
    \toprule
    \textbf{Key event} & \textbf{Event type} & \textbf{Event content} \\
    \midrule
    \endfirsthead
    
    \multicolumn{3}{c}%
    {{\bfseries \tablename\ \thetable{} -- continued from previous page}} \\
    \toprule
    \textbf{Key event} & \textbf{Event type} & \textbf{Event content} \\
    \midrule
    \endhead
    
    \midrule
    \multicolumn{3}{r}{{Continued on next page}} \\
    \bottomrule
    \endfoot
    
    \bottomrule
    \endlastfoot
    
    
    key1 & History & 
    \textbf{Initial medical record:} ...Patient, Female, 71 years old... Admission Diagnosis: 1. Valvular Heart Disease, Severe Mitral Regurgitation, Cardiac Function Class III (NYHA); 2. Coronary Artery Disease, Post-CAG+PCI......Grade 3/6 systolic murmur at the apex...
    \newline \newline
    ...History of Coronary Heart Disease for $>9$ years, oral Aspirin... Clopidogrel... Underwent first coronary stent implantation $>9$ years ago at [Hospital Name Omitted]... underwent coronary stent implantation in [Year]... underwent CAG+PCI surgery on [Year]... Left Anterior Descending branch... severe calcification... in-stent intimal hyperplasia... Right Coronary Artery... implanted Firebird2 $3.5 \times 18$ mm stent... 
    \newline \newline
    ...Echocardiography: Severe Mitral Regurgitation... Left atrial and ventricular enlargement... Aortic valve calcification... Left ventricular diastolic function decreased... [Previous] Echocardiography indicated: Massive eccentric mitral regurgitation... Left heart enlargement...
    \\ 
    \midrule
    
    key2 & Surgery & 
    Mitral Valve Bio-prosthesis Replacement (Preserving Subvalvular Apparatus)
    \\ 
    \midrule
    
    key3 & Clinical Order & 
    Pathogen DNA Detection (KPC)... Pathogen DNA Detection (NDM)... Pathogen DNA Detection (IMP)... Pathogen DNA Detection (VIM)... Pathogen DNA Detection (OXA-48)... Blood Cell Analysis (Three-part differential or above) Determination of C-Reactive Protein (CRP)... Determination of Thrombin Time (TT)... Determination of Plasma Prothrombin Time (PT) Determination of Activated Partial Thromboplastin Time (APTT) Determination of Plasma Fibrinogen Determination of Plasma Antithrombin Activity (ATA) Determination of Fibrin(ogen) Degradation Products (FDP) Determination of Plasma D-Dimer Determination of Thrombin-Antithrombin Complex (TAT) Detection of Plasma Thrombomodulin Antigen (TMAg) Determination of Plasma Plasmin-$\alpha$2-plasmin Inhibitor Complex (PAP) Detection of Plasma Tissue Plasminogen Activator Inhibitor Antigen (t-PAI)... Detection of Procalcitonin (Chemiluminescence) Determination of Interleukin-6 (IL-6)...
    \\ 
    \midrule
    
    key4 & Lab Test & 
        ...Red Blood Cell Count: 2.4 $\times$ $10^{12}$/L (Reference: 3.8--5.1), Low... Hemoglobin: 71 g/L (Reference: 115--150), Low... Procalcitonin: 0.683 ng/ml (Reference: ... $>2$ High risk sepsis), Elevated... Interleukin-6: 261 pg/ml (Reference: 7.0), Elevated... International Normalized Ratio (cobas): 1.16 (Reference: 0.91--1.15), Elevated... Lactate Dehydrogenase: 415 U/L (Reference: 120--246), Elevated... Antithrombin (cobas): 55.5 (Reference: 79.0--113.2), Low... Prothrombin Time (cobas): 11.4 Seconds (Reference: 7.67--10.50), Elevated... Activated Partial Thromboplastin Time (cobas): 56.1 Seconds (Reference: 26.8--42.3), Elevated... Fibrin(ogen) Degradation Products (werfen): 40.6 mg/L (Reference: 0--5), Elevated... D-Dimer (werfen): 7.46 mg/L FEU (Reference: 0--0.5), Elevated... Plasmin-$\alpha$2-Plasmin Inhibitor Complex: 6.123 $\mu$g/mL (Reference: $<0.85$), Elevated... Platelet Count: 109 $\times$ $10^{9}$/L (Reference: 125--350), Low... Thrombin-Antithrombin III Complex: 11.3 ng/mL (Reference: $<4.07$), Elevated... High-Sensitivity C-Reactive Protein: 72.86 mg/L (Reference: 0--6), Elevated... Serum Amyloid A: $>350.00$ mg/L (Reference: 0--10), Elevated...
        \\ 
        \midrule
        
    key5 & Lab Test & 
        ...Neutrophil Count: 10.14 $\times$ $10^{9}$/L (Reference: 1.8--6.3), Elevated... Procalcitonin: 0.594 ng/ml (Reference: 0--0.046 Normal 0.047--0.50 Indicates low risk sepsis or septic shock $>2$ High risk sepsis), Elevated... Interleukin-6: 66.7 pg/ml (Reference: 7.0), Elevated... Hemoglobin: 72 g/L (Reference: 115--150), Low... Fibrinogen cobas: 4.4 g/L (Reference: 2.08--3.85), Elevated... Activated Partial Thromboplastin Time cobas: 62.5 Seconds (Reference: 26.8--42.3), Elevated... Platelet Count: 98 $\times$ $10^{9}$/L (Reference: 125--350), Low... Antithrombin cobas: 52.3 (Reference: 79.0--113.2), Low... Prothrombin Time cobas: 16.6 Seconds (Reference: 7.67--10.50), Elevated... International Normalized Ratio cobas: 1.66 (Reference: 0.91--1.15), Elevated... D-Dimer cobas: 1.19 mg/L FEU (Reference: 0--0.510), Elevated... White Blood Cell Count: 11.52 $\times$ $10^{9}$/L (Reference: 3.5--9.5), Elevated... Thrombin-Antithrombin III Complex: 5.81 ng/mL (Reference: $<4.07$), Elevated...
        \\ 
        \midrule

    key6 & Radiology Report & 
    Abdominal Aortography (Abdominal CTA) (64-slice Contrast-Enhanced CT) ...The abdominal aorta and its main visceral branches are visualized... Bilateral renal arteries, iliac arteries, and the celiac trunk show normal course and morphology... The lumens of various vessels show uneven caliber; vessel walls are thickened with visible plaque-like calcifications... A local filling defect is observed in the superior mesenteric artery... Diagnostic Conclusion: Atherosclerotic changes of the abdominal aorta... Local filling defect in the superior mesenteric artery, considering thrombosis... Please correlate with clinical findings...
    \\ 
\end{longtable}
\end{small}

\begin{figure*}[t]
    \centering
    \includegraphics[width=0.6\linewidth]{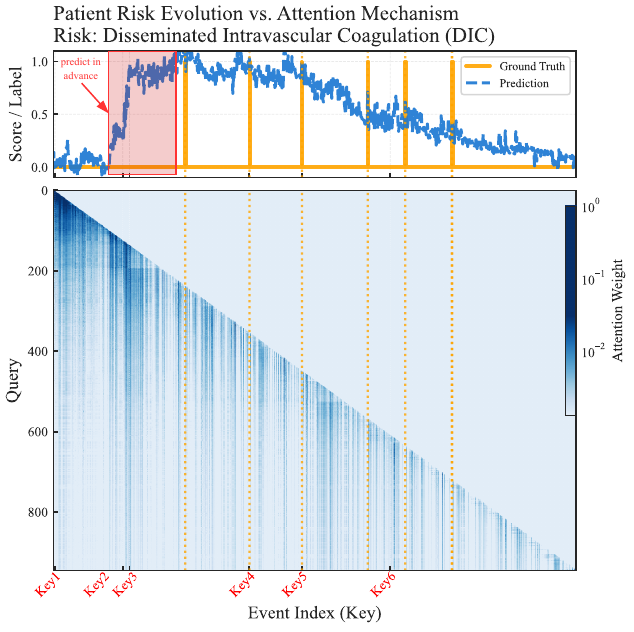}
    \caption{Case Study on Interpretability and Early Prediction (DIC). The top panel tracks the predicted risk score (blue) against ground truth labels (orange), highlighting a successful pre-warning window (red box). The bottom panel displays the attention weight distribution, where distinct vertical high-attention bands indicate the model's mechanism of ``locking'' onto critical clinical events (Key1--Key6) and sustaining this attention over long durations. This verifies the model's capability to capture non-contiguous, sparse dependencies in irregular clinical trajectories.}
    \label{fig:Key_event_hotmap}
\end{figure*}

\subsection{Case Analysis}
\label{case ana}

We conduct a qualitative analysis on a randomly selected patient record. 
Given the complex interplay of multiple co-occurring risks within a single patient trajectory, we isolate \textit{Disseminated Intravascular Coagulation} (DIC) for this specific analysis. 
\Cref{fig:Key_event_hotmap} presents the temporal attention heatmap, the risk intervals manually annotated by medical experts, and the model's predicted risk probability curve. 
Since our model performs simultaneous prediction over 360 distinct clinical risks, the raw sequence contains numerous key events unrelated to the target condition. 
Consequently, for the sake of brevity, \cref{tab:Key_events} exclusively lists the key events relevant to DIC.

The subject of this case study is an elderly female patient with a history of severe cardiovascular conditions, including severe mitral regurgitation, heart failure, and coronary artery disease status post-multiple PCI procedures. 
Against this backdrop, the model leveraged continuous, multi-timepoint objective data to identify risk events that were both logically consistent and clinically verifiable.
The patient underwent ``Mitral Valve Bio-prosthesis Replacement,'' a surgical procedure recognized as a significant risk factor for Disseminated Intravascular Coagulation (DIC). 
Post-operative protocols included a comprehensive panel of advanced molecular biomarkers for early DIC detection. 
From the early stages, the patient exhibited significant elevations in D-Dimer and Fibrin Degradation Products (FDP), prolonged coagulation times, and thrombocytopenia. 
These markers persisted or worsened in subsequent tests, aligning with the typical clinical manifestations of coagulation activation, hyperfibrinolysis, and impaired anticoagulation. 
This pattern indicated a high-risk state of co-existing thrombosis and hemorrhage. 
Crucially, subsequent imaging revealed a local filling defect in the \textit{superior mesenteric artery}, suggestive of thrombosis. 
This finding corroborated the model's prior risk identification, forming a closed logical loop between risk warning and outcome verification. 
This demonstrates the medical plausibility and verifiability of the model's outputs, as it successfully anticipated real risks consistent with subsequent objective findings.

Regarding early warning capabilities, the model placed significant attention on admission information (Key 1). 
Furthermore, by attending to the surgery event (which lacked a specific risk label) and post-operative examination events (Key 2, Key 3), the model successfully issued an early warning for DIC risk. 
Notably, the model maintained an attention distribution aligned with human experts regarding the key events diagnostic of DIC. Crucially, it is the integration of PSL that enables the model to identify these risk-labeled antecedent events prior to clinical onset. By dynamically elevating the predicted probability based on these early signals, our approach achieves anticipatory warning capabilities that are unattainable by conventional binary classification methods.

At the data representation level, by embedding structured data, the model effectively utilized sparse and diverse clinical features for information compression. 
This approach mitigates issues related to information loss or the curse of dimensionality often associated with sparse structured data. 
Compared to ad-hoc manual feature extraction, this embedding-based method is more practical and facilitates smoother deployment in clinical environments.

In summary, the risk events identified by the model in this case are supported by definitive objective evidence and present an evolutionary chain consistent with clinical logic. 
The key risk events were further corroborated by imaging evidence. 
This indicates that the model's risk assessments are not only valid but also possess high clinical interpretability. 
Consequently, the model demonstrates significant value for clinical application, potentially assisting ICU physicians in early risk recognition and intervention.

\section{MIMIC setup}
\label{MIMIC_setup}

\begin{table}[htbp]
\centering
\caption{Cohort selection criteria and experimental design for the three downstream clinical prediction tasks derived from the MIMIC-IV database.}
\label{tab:cohort_selection}

\setlength{\tabcolsep}{4pt}
\begin{tabular}{@{}lccc@{}}
\toprule
\textbf{Criterion} & \textbf{Sepsis Prediction} & \textbf{In-Hospital Mortality} & \textbf{IMV Prediction} \\ \midrule
\textbf{Target Definition} & Sepsis-3 (SOFA shift) & Survival Status (0/1) & First Intubation \\
\textbf{Inclusion Criteria} & Age $\ge$ 18, History $\ge$ 30h & Age $\ge$ 18, LOS $\ge$ 24h & Age $\ge$ 18, Event $>36$h \\
\textbf{Obs. Window} & 24h ($T_{onset}-30$h to $-6$h) & First 24h of Admission & 24h ($T_{onset}-30$h to $-6$h) \\
\textbf{Prediction Gap} & 6 hours & Remainder of Stay & 6 hours \\
\textbf{Control Sampling} & Pseudo-onset($T_{pseudo}$)  & All non-survivors & Pseudo-onset($T_{pseudo}$)  \\
\textbf{positive samples} & 1822 & 7557 &  1999 \\
\textbf{negative samples} & 17857 & 47594 & 23489 \\ \bottomrule
\end{tabular}
\end{table}

Table \ref{tab:cohort_selection} outlines the detailed sample selection criteria and experimental design for the three downstream clinical prediction tasks. To ensure rigorous evaluation, we adopted a unified multimodal feature extraction strategy while tailoring the temporal windowing to the specific clinical nature of each risk. For the Sepsis and Invasive Mechanical Ventilation (IMV) tasks, we implemented a dynamic prediction paradigm: positive samples are anchored to the specific event onset ($T_{onset}$), while negative controls utilize randomly sampled pseudo-onset times ($T_{pseudo}$) within valid observation intervals. A strict prediction gap is enforced to simulate realistic early warning scenarios, ensuring the model relies solely on historical data without leakage from the imminent pre-event phase. Conversely, the In-Hospital Mortality task employs a static early-observation design, utilizing data exclusively from the first 24 hours of ICU admission to prognosticate the final survival outcome. This distinction allows us to evaluate the model's versatility in handling both event-triggered dynamic monitoring and static admission-based risk stratification. Crucially, to ensure reproducibility and alignment with community benchmarks, all event onset determinations are strictly derived from the standardized SQL query scripts provided in the official MIT-LCP Code Repository\footnote{\url{https://github.com/MIT-LCP/mimic-code}}.

Furthermore, consistent with the Unified Semantic Embedding framework proposed in the main text, all multi-modal features extracted from the MIMIC-IV cohorts underwent the identical serialization and encoding process. Specifically, heterogeneous inputs—ranging from high-frequency tabular vitals to unstructured nursing notes—were converted into the standardized text stream format and projected into the semantic space via the pre-trained embedding model, ensuring strict feature alignment across all experimental tasks. However, distinct from the SIICU protocol, these experiments did not employ the Plateau-Gaussian Soft Labeling (PSL) mapping. Instead, we formulated the prediction of each risk as a standard binary classification task, optimized using the Binary Cross-Entropy (BCE) loss.

To ensure a rigorous assessment of model generalization, we adopted a patient-level random splitting protocol with an $8:1:1$ distribution for training, validation, and testing, respectively. All experiments were repeated across four independent iterations using distinct random seeds. The final results are presented as the mean and standard deviation of these four runs, ensuring that the performance gains are statistically significant and not artifacts of a specific data split.

\section{Supplementary Experiment}
\label{sec:metric_stability}
\subsection{Stability}

\begin{figure*}[t]
    \centering
    
    \includegraphics[width=0.45\linewidth]{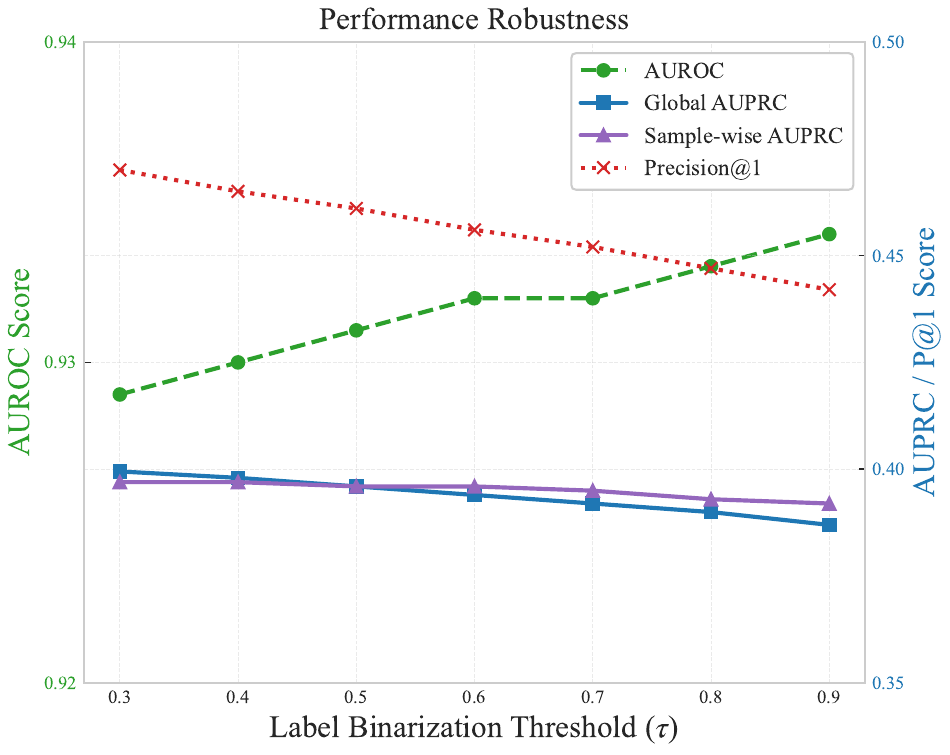}
    \caption{Sensitivity Analysis of Label Binarization Threshold ($\tau$). The figure illustrates the model's performance stability across a wide range of risk severity definitions ($\tau \in [0.3, 0.9]$). While Precision@1 naturally decays as the threshold broadens, the ranking metrics (Global and Sample-wise AUPRC) exhibit negligible fluctuation ($\Delta < 0.013$). This \textit{threshold invariance} confirms that MATA-Former robustly captures the continuous progression of risk rather than overfitting to a specific binary cutoff.}
    \label{fig:threshold_perform} 
    
    \vspace{0.2in} 
    
    \includegraphics[width=\linewidth]{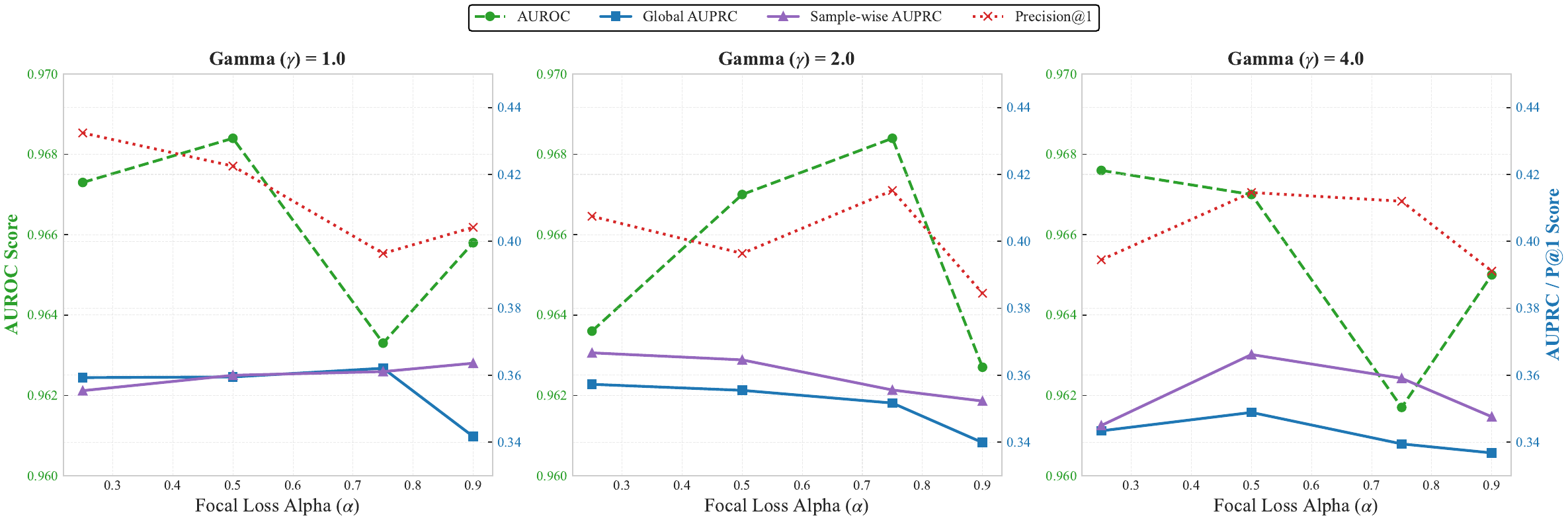}
    \caption{Ablation Study on Focal Loss Hyperparameters. To investigate the efficacy of handling zero-inflated data via Focal Loss, we conducted a grid search over the focusing parameter $\gamma \in \{1.0, 2.0, 4.0\}$ and the balancing factor $\alpha$. The results demonstrate significant performance volatility and a failure to consistently surpass the unweighted MSE baseline across various configurations. This empirical evidence supports the adoption of unweighted MSE as the superior objective function for learning continuous clinical risk trajectories.}
    \label{fig:focalloss_ablation} 
    
\end{figure*}

Although the model is optimized as a multi-task regression framework to fit continuous risk trajectories, assessing its clinical utility necessitates evaluating the capacity to discriminate between high-risk and low-risk states. Consequently, we map the continuous ground-truth labels back into a binary classification format for metric computation. We define a label binarization threshold, denoted as $\beta$, to reconstruct the binary ground truth $Y_{n}^{(r,h), \text{bin}}$ from the continuous soft labels $Y_{n}^{(r,h)}$:
\begin{equation}
    Y_{n}^{(r,h), \text{bin}} = \mathbb{I}(Y_{n}^{(r,h)} > \beta)
\end{equation}
where $\mathbb{I}(\cdot)$ denotes the indicator function. This protocol enables the employment of standard ranking metrics for evaluation while retaining the benefits of continuous supervision inherent to the training phase.

Since the selection of $\beta$ implicitly defines the severity scope of risk events—where a lower $\beta$ incorporates the early rise phase as positive, while a higher $\tau$ focuses strictly on the sustained critical phase—we conducted a sensitivity analysis to verify the robustness of MATA-Former. As illustrated in \cref{fig:threshold_perform}, we evaluated the ranking performance (AUPRC) across a threshold range of $\beta \in [0.3, 0.9]$. The results exhibit negligible performance fluctuation ($\Delta \text{AUPRC} < 0.013$). This threshold invariance confirms that:
\begin{enumerate}
    \item The model successfully identifies high-risk events rather than merely overfitting to the distributional tails.
    \item Regardless of whether the risk definition is broad ($\beta=0.3$) or conservative ($\beta=0.9$), the model consistently assigns higher prediction scores to true positive cases, validating that MATA-Former has captured the latent continuity of disease progression.
\end{enumerate}

\subsection{Focal loss}

To address the prevalence of zero-inflated samples in our dataset, we evaluated Focal Loss, which is widely regarded as effective for class imbalance. Formally, for a soft target $y \in [0, 1]$ and the sigmoid-activated prediction $\hat{y} \in (0, 1)$, we adapted the objective as:
\begin{equation}
    \mathcal{L}_{\text{Focal}} = - \left[ \alpha (1 - \hat{y})^\gamma y \ln \hat{y} + (1 - \alpha) \hat{y}^\gamma (1 - y) \ln (1 - \hat{y}) \right]
\end{equation}
where $\alpha$ balances positive/negative importance and $\gamma$ controls the focusing effect on hard examples. However, contrary to the general consensus, our empirical results indicate that Focal Loss underperforms compared to the standard unweighted MSE in this specific regression context. To verify that this discrepancy is not attributable to suboptimal hyperparameter configuration, we conducted a rigorous sensitivity analysis of Focal Loss while keeping the model architecture and data fixed. As illustrated in \cref{fig:focalloss_ablation}, we systematically varied the focusing parameter $\gamma$ and the balancing factor $\alpha$. The results demonstrate that, across a wide range of parameter settings, the performance of Focal Loss consistently fails to surpass the MSE baseline. This finding substantiates that the unweighted MSE objective is more robust for learning the continuous risk trajectories in our high-sparsity clinical data.

\subsection{Semantic Manifold and Time-Aware Encoding}
\label{Time Encoding}
To embed continuous temporal dynamics into the high-dimensional semantic space, we define a mapping function $\Phi: \mathbb{R}^+ \rightarrow \mathbb{R}^{d_{model}}$. Given a physical timestamp $t$, the encoding for the $k$-th dimension is formulated as:

\begin{equation}
    \Phi(t, k) = 
    \begin{cases} 
    \sin\left( \omega_j \cdot t \right), & \text{if } k = 2j \\
    \cos\left( \omega_j \cdot t \right), & \text{if } k = 2j+1 
    \end{cases}
\end{equation}

\noindent where the angular frequency $\omega_j$ decays geometrically with the dimension index $j \in [0, d_{model}/2 - 1]$:

\begin{equation}
    \omega_j = \frac{1}{T^{2j / d_{model}}}
\end{equation}

\noindent Here, $T=1209600.0$ serves as the temporal scaling horizon.

A structural conflict exists between the pre-trained semantic space and the temporal projection. The input embeddings $\mathbf{x}$ from the Qwen model occupy a unit hypersphere, i.e., $\|\mathbf{x}\|_2 = 1$. In contrast, the $L_2$-norm of the positional encoding $\mathbf{p} = \Phi(t)$ is significantly larger:

\begin{equation}
    \|\mathbf{p}\|_2 = \sqrt{\sum_{k=0}^{d_{model}-1} \Phi(t, k)^2} = \sqrt{\sum_{j=0}^{\frac{d_{model}}{2}-1} \left(\sin^2(\dots) + \cos^2(\dots)\right)} = \sqrt{\frac{d_{model}}{2}}
\end{equation}

\noindent For $d_{model}=4096$, this yields $\|\mathbf{p}\|_2 \approx 45.25$. Direct addition would result in a signal-to-noise ratio of $1:45.25$, causing \textit{semantic collapse}. To mitigate this, we scale the embeddings by a factor $\gamma = 64$:

\begin{equation}
    \mathbf{z} = \gamma \cdot \mathbf{x} + \mathbf{p}, \quad \text{where } \gamma \approx \sqrt{d_{model}}
\end{equation}

\noindent This adjustment restores the signal magnitude ($64$) to be comparable to the temporal noise ($45.25$), maintaining a ratio of $\approx 1.4:1$. Furthermore, aligning the magnitude with $\sqrt{d_{model}}$ optimizes the gradient flow within the subsequent Self-Attention mechanism, defined as $\text{Softmax}(\frac{QK^T}{\sqrt{d_{model}}})$.

\end{document}